\documentclass[twoside]{article}
\usepackage[a4paper]{geometry}
\usepackage[utf8]{inputenc}
\usepackage[T1]{fontenc} 
\usepackage{RR}
\usepackage{subfig}
\usepackage[inline]{enumitem}
\usepackage{hyperref}

\usepackage{amsmath}
\usepackage{amssymb}
\usepackage{mathtools}
\newtheorem{theorem}{Theorem}
\newtheorem{proposition}{Proposition}

\newcommand{\SC}{{\cal S}}
\newcommand{\A}{{\cal A}}
\RRNo{9503}
\RRdate{April 2023}
\RRauthor{
Hector Kohler
  \thanks[sfn]{Univ. Lille, CNRS, Inria, Centrale Lille, UMR 9189 – CRIStAL, Lille, France}%
  \and
Riad Akrour\thanks{Univ. Lille, Inria, CNRS, Centrale Lille, UMR 9189 – CRIStAL, Lille, France}%
\and Philippe Preux\thanksref{sfn}}
\authorhead{H. Kohler et al.}
\RRtitle{Compromis Interpr\'etabilit\'e-Performance Optimale des Arbres de Classification avec de l'Apprentissage par Renforcement Bo\^ite Noire}
\RRetitle{Optimal Interpretability-Performance Trade-off of Classification Trees with Black-Box Reinforcement Learning}
\titlehead{Classification Trees with RL}
\RRresume{L'interpr\'etabilit\'e des mod\`eles d'IA permet \`a l'utilisateur d'effectuer des contr\^oles de s\'ecurit\'e afin d'instaurer la confiance dans ces mod\`eles. En particulier, les arbres de d\'ecision fournissent une vue d'ensemble du mod\`ele appris et soulignent clairement le r\^ole des caract\'eristiques essentielles à la classification de ces donn\'ees. Cependant, l'interpr\'etabilit\'e est entrav\'ee si l'arbre de d\'ecision est trop grand. Pour apprendre des arbres compacts, un cadre d'apprentissage par renforcement a \'et\'e r\'ecemment propos\'e pour explorer l'espace des arbres de d\'ecision. Une t\^ache de classification supervis\'ee donn\'ee est mod\'elis\'ee comme un probl\`eme de d\'ecision de Markov auquel on ajoute des actions qui r\'ecoltent des informations sur les caract\'eristiques d'une donn\'ee, ce qui \'equivaut \`a la construction d'un arbre de d\'ecision. En p\'enalisant ces actions de mani\`ere appropri\'ee, l'agent RL apprend \`a faire un compromis optimal entre la taille et les performances de l'arbre appris. 
Cependant, un agent doit r\'esoudre un probl\`eme de d\'ecision de Markov partiellement observable. 
La principale contribution de cet article est de prouver qu'il suffit de r\'esoudre un probl\`eme enti\`erement observable pour apprendre un arbre de d\'ecision optimisant le compromis interpr\'etabilit\'e-performance. Ainsi, n'importe quel algorithme de planification ou d'apprentissage par renforcement peut \^etre utilis\'e. Nous d\'emontrons l'efficacit\'e de cette approche sur un ensemble de t\^aches de classification supervis\'ee et nous la comparons \`a d'autres m\'ethodes d'optimisation de l'interpr\'etabilit\'e et de la performance.}
\RRabstract{Interpretability of AI models allows for user safety checks to build trust in these models. In particular, decision trees (DTs) provide a global view on the learned model and clearly outlines the role of the features that are critical to classify a given data. However, interpretability is hindered if the DT is too large. To learn compact trees, a Reinforcement Learning (RL) framework has been recently proposed to explore the space of DTs. A given supervised classification task is modeled as a Markov decision problem (MDP) and then augmented with additional actions that gather information about the features, equivalent to building a DT. By appropriately penalizing these actions, the RL agent learns to optimally trade-off size and performance of a DT. 
However, to do so, this RL agent has to solve a partially observable MDP. 
The main contribution of this paper is to prove that it is sufficient to solve a fully observable problem to learn a DT optimizing the interpretability-performance trade-off. As such any planning or RL algorithm can be used. We demonstrate the effectiveness of this approach on a set of classical supervised classification datasets and compare our approach with other interpretability-performance optimizing methods.}
\RRmotcle{Apprentissage par Renforcement, Apprentissage Supervis\'e, Interpr\'etabilit\'e}
\RRkeyword{Reinforcement Learning, Supervised Learning, 
 Interpretability}
\RRprojet{Scool}  
\RCLille 

\begin{document}
\makeRR 
\section{Introduction}
The last decade or so has seen a surge in the performance of machine learning models, whether in supervised learning~\cite{AlexNet} or RL ~\cite{mnih2015human}. These achievements rely on deep neural models that are often described as black-box~\cite{Murdoch,Guidotti18,Arrieta}, trading interpretability for performance. In many real world tasks, predictive models can hide undesirable biases~(see e.g. Sec. 2 in \cite{Guidotti18} for a list of such occurrences) hindering trustworthiness towards AIs. Gaining trust is one of the primary goals of interpretability~(see Sec.\@ 2.4 of \cite{Arrieta} for a literature review) along with informativeness requests, i.e.\@ the ability for a model to provide information on why a given decision was taken. The computational complexity of such informativeness requests can be measured objectively, and \cite{Barcelo20} showed that multi-layer neural networks cannot answer these requests in polynomial time, whereas several of those are in polynomial time for linear models and DT.

In contrast to deep neural models, DTs provide a global look at the learned model and transparently reveal which features of the input are used in taking a particular decision. This is referred to as global \cite{Guidotti18} or model-based \cite{Murdoch} interpretability, as opposed to post-hoc interpretability \cite{Murdoch,Arrieta}. Even though DTs are globally intepretable, they have also been used in prior work for post-hoc interpretability of deep neural models, e.g.\@ in image classification \cite{Zhang19} or RL \cite{Viper}. The latter work provides another motivation for DTs, as their simpler nature allowed to make a stability analysis of the resulting controllers and provided theoretical guarantees of their efficacy. In the 54 papers reviewed in \cite{Guidotti18}, over $25\%$ use DTs as the interpretable model and over $50\%$ the more general class of decision rules.

DTs are a common interpretable model and it is thus important to improve 
their associated learning algorithms. However, interpretability of DTs is hindered if the tree grows too large. The quantification of what is too large might vary greatly depending on the desired type of simulability, that is whether we want individual paths from root to leaf to be short or the total size of the tree to be small \cite[p.\@ 13]{Lipton}. In both cases, an algorithmic mechanism to control these tree metrics and to manage the inevitable trade-off between interpretability and performance is necessary. One of the main challenges for learning DTs is that it is a discrete optimization problem that cannot, a priori, be solved via gradient descent. Algorithms such as CART \cite{Cart} build a DT by greedily maximizing the information gain---a performance related criteria. Interpretability can then be controlled by fixing a maximal tree depth, or by using post-processing pruning algorithms \cite{Bradford,Prodromidis}. Unfortunately, this two-step process provides no guaranty that the resulting DT is achieving an optimal interpretability-performance trade-off.

An alternative way to learn DTs, that inherently takes into account the interpretability-performance trade-off, is the recently proposed framework of Iterative Bounding Markov Decision Processes (IBMDPs) \cite{IBMDP}. An IBMDP extends a base MDP state space with \textit{feature bounds} that encode the current knowledge about the input, and the action space with \textit{information gathering actions} that refine the feature bounds by performing the same test a DT would do: comparing a feature value to a threshold. The reward function is also augmented with a penalty to take the cost of information gathering action into account. The IBMDP reward function encodes an interpretability-performance trade-off: an agent learns when to add decision nodes or when to make a prediction. 

In this work we study the IBMDP setting when the base MDPs encode supervised classification tasks. By doing so, we are able to analyse the optimality of policies learned with RL with respect to the IBDMP reward function. Thus our work studies RL frameworks to learn DTs that trade-off between interpretability (depth of the DT) and performance (accuracy of the DT). After a literature review and the introduction of our notations in Section 2 and 3, our work continues as follows:\\
\textbf{Section 4}: we present a simple toy task to benchmark RL algorithms solving IBMDPs and analyse causes of failure of existing RL algorithms solving IBMDPs. \\
\textbf{Section 5}: we present a new RL framework with optimality guarantees w.r.t the IBDMP objective. \\
\textbf{Section 6}: we apply this framework to UCI \cite{UCI} supervised classification datasets. 

All proofs are provided in Appendix. All learned DTs and the code to reproduce all experiments can be found on an \underline{anonymous github}\footnote{ https://github.com/KohlerHECTOR/Interpretability-Performance-official-implem \label{git}}.

\section{Decision Trees for Supervised Learning}
\subsection{Greedy Approaches}
Early research on the induction of DTs focused on the ID3 algorithm, which uses a greedy approach to select the best attribute at each node based on information gain \cite{ID3}. This approach was later extended by the C4.5 algorithm, which introduced techniques such as pruning and handling missing data \cite{C45}. 

However, these methods may lead to large trees that a human cannot interpret.
An other very well-known DT induction algorithm is CART \cite{Cart} which performs equivalently to C4.5 and has the same troubles with the induction of large trees.

\subsection{Optimal Decision Trees}
The algorithms discussed earlier are greedy heuristics and may produce poorly performing trees \cite{kearns1996boosting}. As a result, there is an increasing interest in developing algorithms that can train DTs to achieve optimal accuracy.

Training optimal DTs for classification can done with dynamic programming \cite{nijssen2007mining} or with Mixed-Integer Linear Programming based formulations \cite{bertsimas2017optimal,verwer2017learning}.
However there is no guarantee that the optimal DT will not grow very large. 

To provide regularization and encourage interpretability, the size of the tree is typically limited, and then a solver is used to find the DT that maximizes accuracy within the predetermined size constraints. This cannot be considered as optimizing an interpretability-performance trade-off as the size of the resulting DT is given by the user and not learned. 
\subsection{Online Learning}
Another approach to learn DTs for classification is to model the classification task as an MDP. When doing so, states correspond to training samples, and actions build the DT (either add a decision node or a leaf node by making a prediction). The reward function of the MDP encodes a trade-off between the depth of the tree and the performance of the tree. Indeed, there is a penalty for querying information about a training sample feature, and rewards for predictions. 

The proposed algorithms of \cite{bonet1998learning} and \cite{IBMDP} are RL agents solving Partially Observable MDPs (POMDPs) \cite[chapter 3]{PDMIA}. In \cite{garlapati2015reinforcement}, an other RL agent is proposed, this time acting in a fully observable MDP and is a special case of Iterative Bounding MDPs \cite{IBMDP} where training samples have categorical features. 

None of these works study the optimality of their proposed method with respect to the interpretability-performance trade-off.

\section{Preliminaries}

\subsection{Supervised Classification Tasks}

In this work, we aim to learn DTs for supervised classification tasks. We consider classification tasks made of a set of training examples $\mathcal{X} = \{x_1, ..., x_N\}$ (a dataset), and a set of labels $\mathcal{Y} \in \{C_1, ..., C_K\}^N$ (one of K classes for each training example). Each of the $N$ training example $x_i \in \mathcal{X}$ has $d$ features $x_{i1}, ..., x_{id}$. The goal of the task is to find a classifier $g:\mathbb{R}^d \rightarrow \mathcal{Y}$, $g: x_i \mapsto \hat{y_i}$. Classifiers are computed by algorithms optimizing a loss function of $\hat{y}_i = g(x_i)$ and the true label $y_i$. In this paper we present algorithms returning classifiers that are DTs optimizing a function of both the performance and the interpretability defined in the following sections.

\subsection{Markov Decision Problems}\label{sec:mdp-def}

We consider an infinite horizon MDP \cite{puterman} defined by the tuple $\langle \SC, \A, R, T, \gamma \rangle$, where $\SC$ is the state space, $\A$ is the discrete action space, $R: \SC \times \A \mapsto [R_{\min}, R_{\max}] \subset \mathbb{R}$ is the reward function, $T$ is the transition function, and $\gamma<1$ is the discount factor. The agent interacts with the environment according to its policy $\pi$. At time $t$, 
the agent takes action ${a}_t \sim \pi(. |{s}_t), \ {a}_t \in {\A}$, after which it observes the reward $r_t$ and the next state ${s}_{t+1}$ with probability $T({s}_t, {a}_t, {s}_{t+1})$. 
Let $Q^\pi(s, a) = \mathbb{E}_{\pi} [\sum_{t\ge{}0} \gamma^t R({s}_t, {a}_t) \mid {s}_0 = s, a_0 = a]$ be the Q-function,  $V^\pi(s)=\mathbb{E}_{\pi}[Q(s,a)]$ be the value function, $A^\pi(s,a) = Q^\pi(s, a) - V^\pi(s)$ be the advantage function, and $J(\pi) = \mathbb{E}[V(s_0)]$ be the policy return for some initial state distribution.
In this work we study RL algorithms that find a policy $\pi^*$ that maximizes $J$.

\subsection{Classification Markov Decision Problems}

Any supervised classification task can be cast into a classification MDP  \\$\langle \mathcal{X}, \{C_1, ..., C_K\}, R, T, \gamma \rangle$. If the dataset to be classified is $\mathcal{X} \subsetneq \mathbb{R}^{N \times d}$ then the state space of the MDP is $\mathcal{X}$, that is the set of training examples. If the set of labels is $\mathcal{Y} \in \{C_1, ..., C_K\}^N$, then the action space is $\{C_1, ..., C_K\}$. The transition function is stochastic, we simply transit to a new state (draw a new data point to classify) whatever the action is: $T(x_i, C_h, x_j) = \frac{1}{N}$. The reward function depends on the current state and action: $R(x_i, C_h) = 1 \text{ if } y_i = C_h$ in the supervised classification task; $R(x_i, C_h) = -1 \text{ otherwise}$. A policy $\pi: x_i \mapsto C_h$ is a classifier, and a 
policy $\pi$ that maximizes the expected discounted cumulative reward, also maximizes the classification accuracy. 

\subsection{Iterative Bounding Markov Decision Problems}\label{sec:ibmdp}

\subsubsection{Definition}

Following \cite{IBMDP}, we introduce the notion of an Iterative Bounding MDP (IBMDP). IBMDPs are MDPs. Let us consider a Classification MDP $\langle \mathcal{X}, \{C_1, ..., C_K\}, R, T, \gamma \rangle$. 
 We assume $\mathcal{X} = [0, 1]^{N\times d}$. An Iterative Bounding MDP $\langle \SC^{\prime}, \A^{\prime}, R^{\prime}, T^{\prime}, \zeta, p, \gamma\rangle$ is defined on top of it with the following properties.

\paragraph{State space} $\SC^{\prime} = \mathcal{X} \times \Omega$, with $\Omega \subsetneq [0,1]^{2d} $. A state $s\in \SC^{\prime}$ has two parts. A training sample $x_i = (x_{i1}, ..., x_{id})\in \mathcal{X}$, and feature bounds $o=(L_{1}, ..., L_{d}, U_1, ..., U_d)  \in \Omega$. $(L_k, U_k)$. For each feature of the training sample $x_{ik}$, $(L_k, U_k)$ represents the current known range of its value. Initially, $(L_k, U_k) = (0, 1)$ for all $k$, which are iteratively refined by taking Information-Gathering  Actions (IGAs) defined below.

\paragraph{Action space} $\A^{\prime} = \{C_1, ..., C_K\} \cup \A_I$. An agent in an IBMDP can either take a base action $a \in \{C_1, ..., C_K\}$, or an IGA in $\A_I = \{1,\dots, d\} \times \{\frac{1}{p+1}, ..., \frac{p}{p+1}\}$, with parameter $p \in \mathbb{N}$.

\paragraph{Transition function.} If $a \in \{C_1, ..., C_K\}$, a new training sample is drawn at random from the state space $\mathcal{X}$, while feature bounds are reset to $(0, 1)$. If $a\in \A_I$, the base state is left unchanged, but the feature bounds are refined. Given a training sample $x_i$ with feature bounds $o = (L_1, ..., L_d, U_1, ..., U_d)$ The information gathering action $a = (k, v)$ will compare $x_{ik}$ to $v' = v \times (U_k - L_k) + L_k$, and will set the lower bound $U_k$ to $v'$ if $x_{ik} > v'$, otherwise $L_k$ is set to $v'$. 

\paragraph{Reward function.}
The reward for a base action in $\{C_1, ..., C_K\}$ is defined by the base classification MDP reward function $R$. For an IGA in $\A_I$ the reward is a fixed value $\zeta \in (-\inf, R_{\max}) \equiv \zeta < 1 $ (the maximum value of the base reward function). We impose $\zeta < 1$, as otherwise a policy never taking any base action would always be optimal, though this restriction is not enough to prevent this degenerate case for RL algorithms.

\paragraph{Objective function.} Solving an IBDMP is finding a polciy $\pi^* \in \Pi: \SC' \rightarrow \A'$ such that $\pi^* = \underset{\pi \in \Pi}{\operatorname{argmax}} J(\pi)$ where $J()$ is the expectation of cumulative discounted rewards given by $R'$ (the MDP objective function of Section \ref{sec:mdp-def}).

\subsubsection{Learning a DT using Partially Observable RL}\label{sec:partial-obj} 

 \begin{figure}
    \centering
    \subfloat[Initialisation of the IBMPD.\label{fig:a}]{%
      \includegraphics[width=0.3\textwidth]{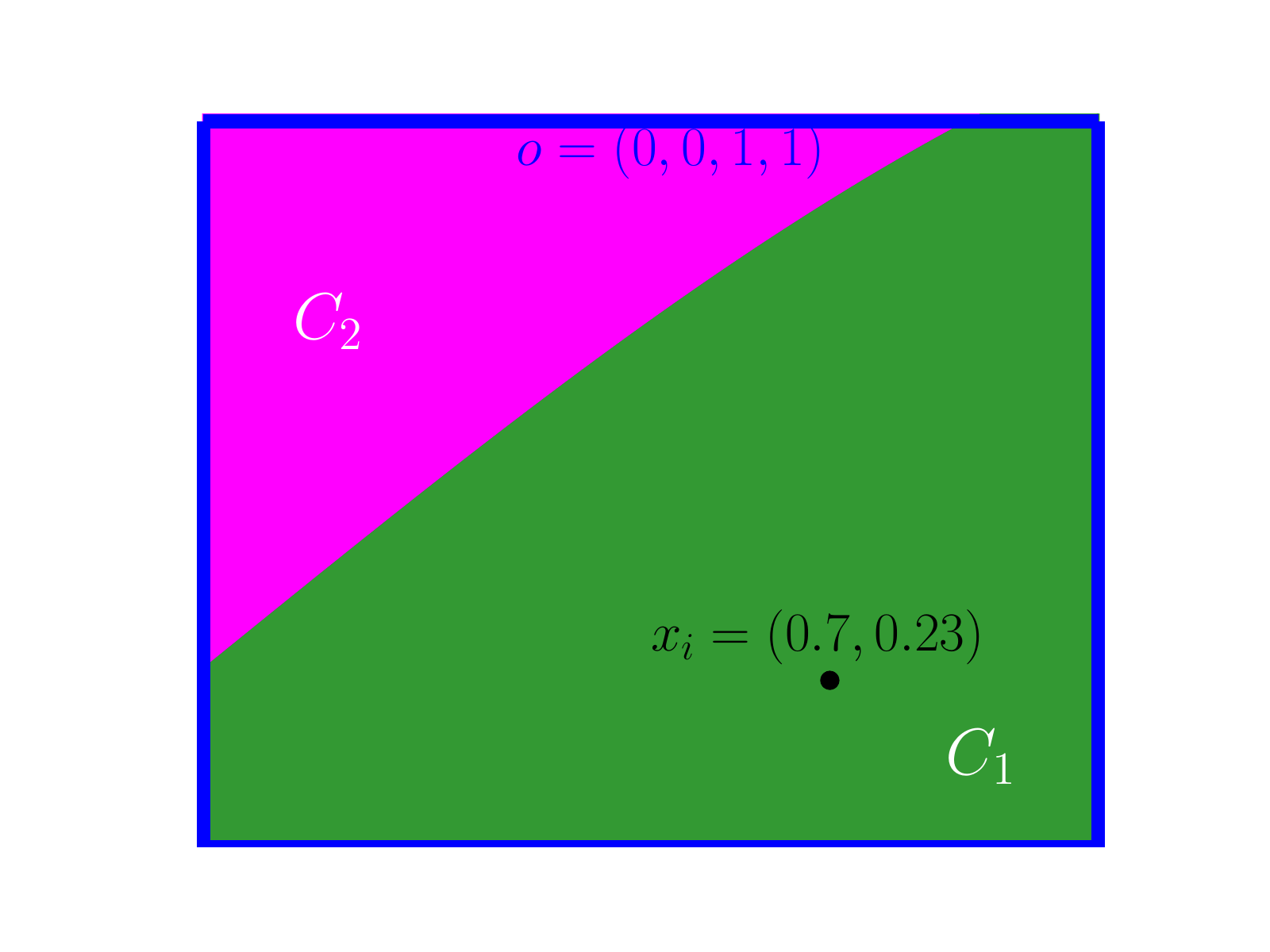}
    }\\
    \vskip -0.05in
    \subfloat[We take IGA $(x_{i2}, 0.5)$ and receive reward $\zeta$.\label{fig:b}]{%
      \includegraphics[width=0.3\textwidth]{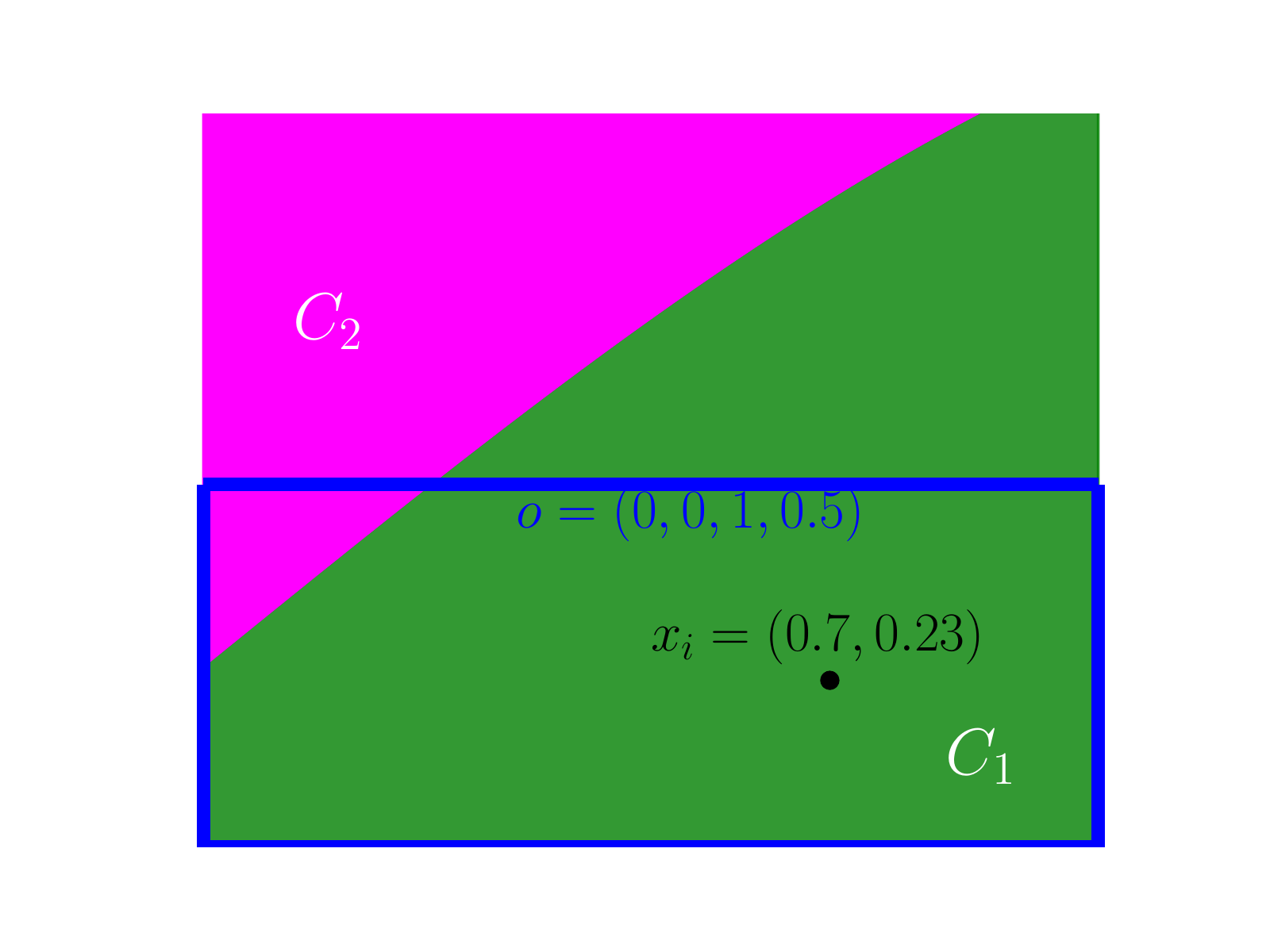}
    }
    \subfloat[We take IGA $(x_{i2}, 0.25)$ and receive reward $\zeta$.\label{fig:d}]{%
      \includegraphics[width=0.3\textwidth]{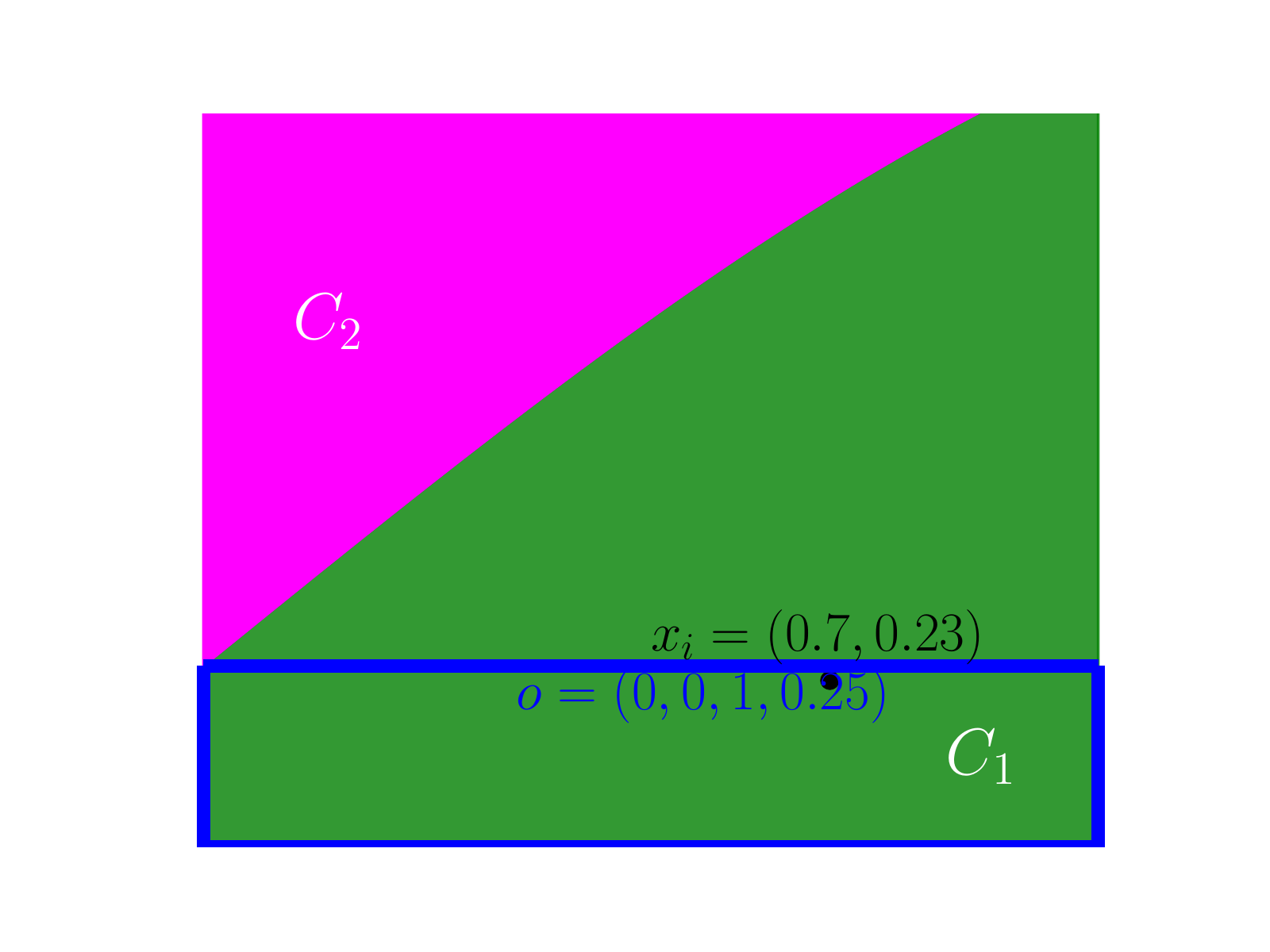}
    }
    \subfloat[We take base action $C_1$ and receive reward 1 as $x \in C_1$\label{fig:f}.]{
      \includegraphics[width=0.3\textwidth]{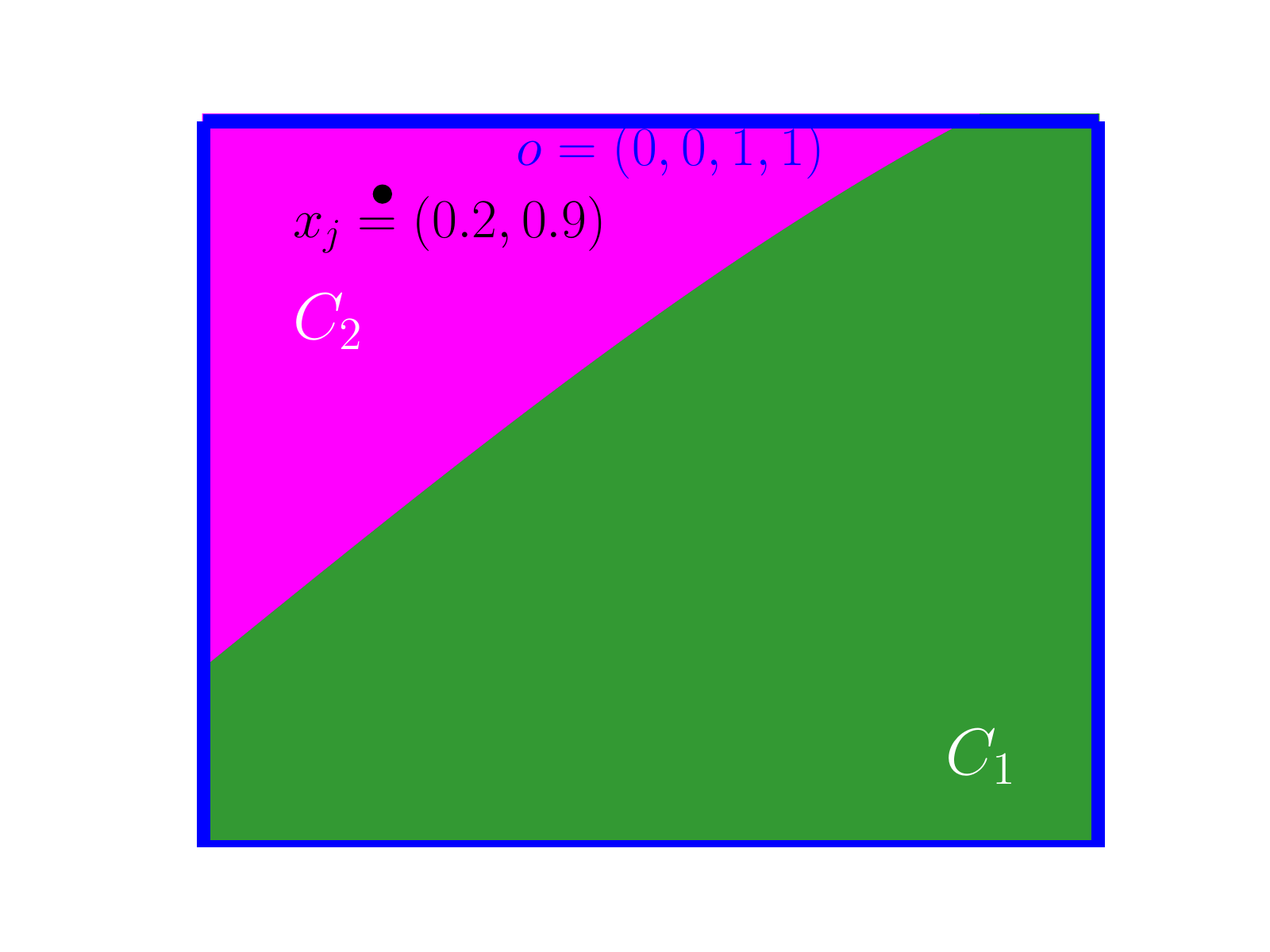}
    }
    \\
        \subfloat[Taking an IGA in the IBMDP adds a decision node to a DT.\label{fig:c}]{%
      \includegraphics[width=0.3\textwidth]{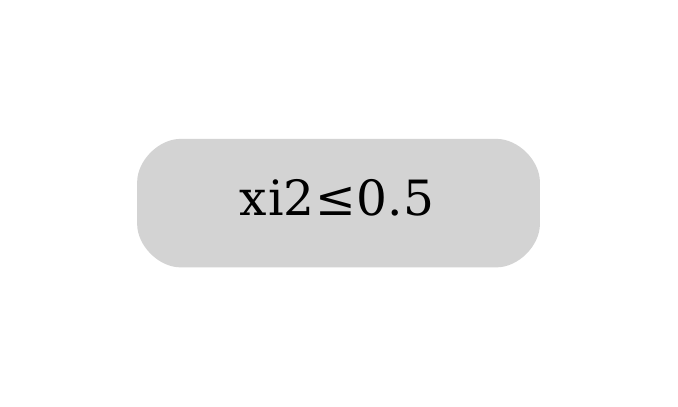}
    }
    \subfloat[Taking an IGA in the IBMDP adds a decision node to a DT.\label{fig:e}]{%
      \includegraphics[width=0.3\textwidth]{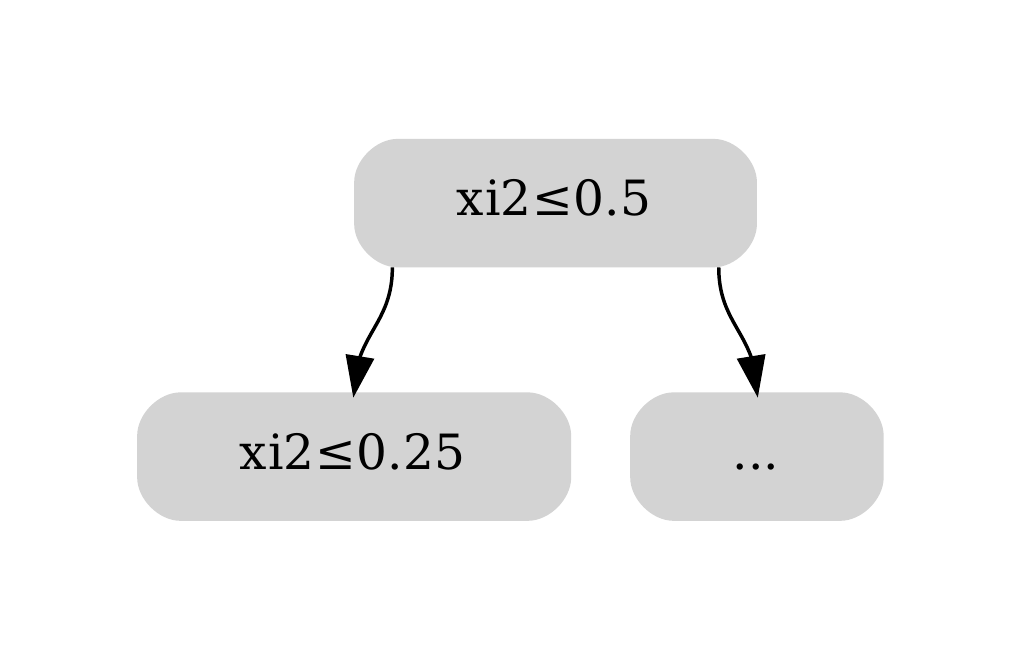}
    }
    \subfloat[Taking a base action add a decision node to the DT.\label{fig:g}]{%
      \includegraphics[width=0.3\textwidth]{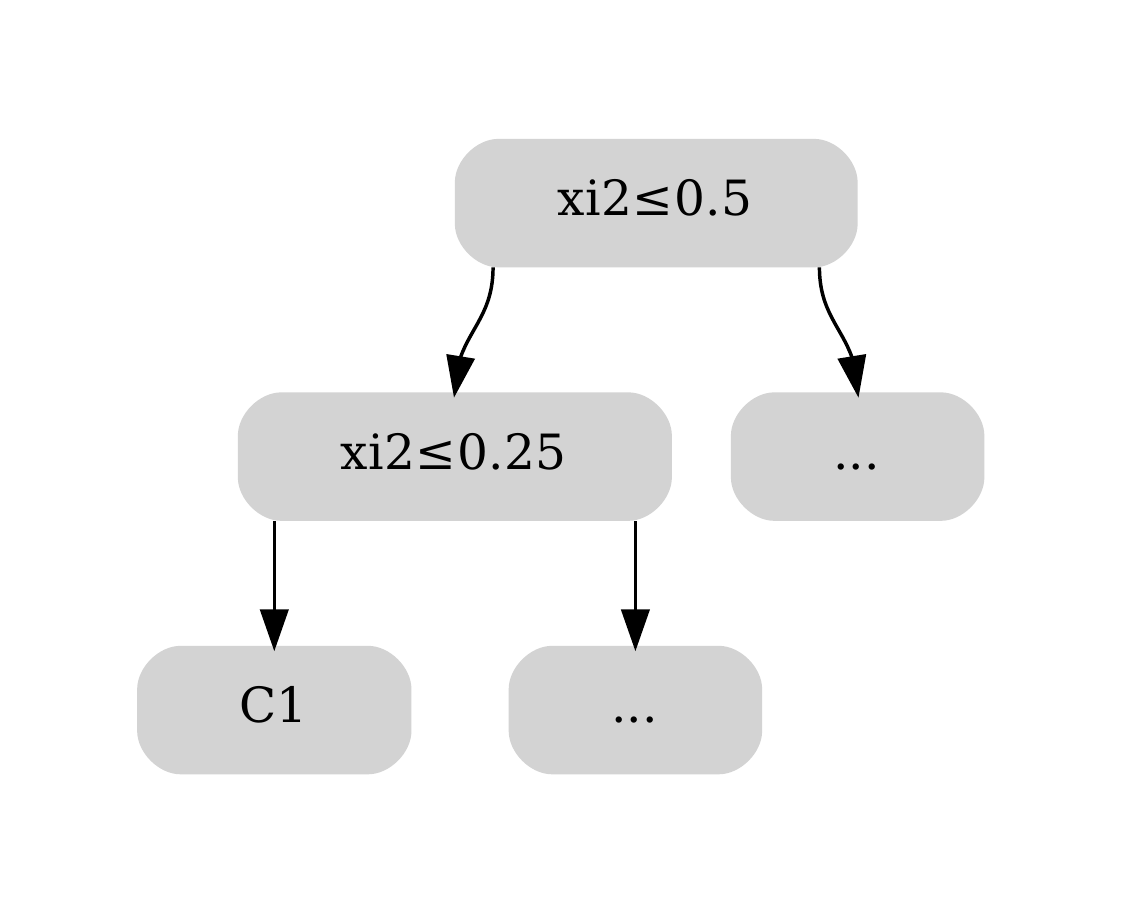}
    }
    \caption{Example trajectory of an IBMDP$\langle p=1\rangle$. The state space is divided in two; green states are training samples with label $C_1$, magenta states are training samples with label $C_2$. (\ref{fig:a}): the IBMDP is initialised: the base state $x_i$ is drawn at random from the base MDP and the feature bounds $o$ are set to $(0,0,1,1)$. (\ref{fig:b}): the agent takes the IGA $(x_{i2}, 0.5)$; the observation part is updated to $ o = (0,0,1,0.5)$ because $x_{i2} = 0.23\leq 0.5$. Another IGA is taken in (\ref{fig:d}). (\ref{fig:f}): the agent takes a base action, so a new base state $x_j$ is drawn from the base transition function and the feature bounds are reset: $o = (0,0,1,1)$.}
    \label{fig:dt-ibmdp1}
\end{figure}

As stated in \cite{IBMDP}, a RL algorithm for an IBMDP should return a policy depending on feature bounds only in order to be able to extract a DT. So an agent learns a DT optimizing an interpretability-performance trade-off encoded by an IBMDP reward function by finding a policy $\pi^* \in \Pi_{DT}: \Omega \rightarrow \A'$ such that $\pi^* = \underset{\pi \in \Pi_{DT}}{\operatorname{argmax}} J(\pi)$  . We illustrate how an agent learning such a policy is equivalent to learning a DT in Figure \ref{fig:dt-ibmdp1}. To learn a policy depending on feature bounds only, \cite{IBMDP} proposes CUSTARD, a partially observable RL algorithm learning a policy depending only on the feature bounds of the IBMDP state and value functions depending on the full IBMDP state. We connect CUSTARD to the class of asymmetric RL algorithms first studied empirically in \cite{Pinto} and more recently theoretically in \cite{Baisero,baisero22a}.

\paragraph{Asymmetric Q-Learning.}\label{cust-dqn} In asymmetric Q-Learning methods, like the DQN \cite{mnih2015human} version of CUSTARD \cite{IBMDP}, an oracle state-action function depending on the full state of the IBMDP is learned with TD-learning \cite{sutton2018reinforcement}. This oracle $Q$-function is used as target for the TD-learning of an other state-action value function, this time, that depends only on feature bounds. 
\paragraph{Asymmetric actor-critic.}\label{cust-ppo} In asymmetric actor-critic methods, like the PPO \cite{Schulman17} version of CUSTARD \cite{IBMDP}, a value function depending on the full state of the IBMDP is learned and a policy depending only on feature bounds is learned. Note that the policy gradient theorem \cite{sutton00} still holds in the asymmetric setting.
We show in the next section that CUSTARD fails to learn DTs for simple tasks.

\section{Partially Observable RL for simple Classification Tasks}
\subsection{A Binary Classification Benchmark}\label{sec:xptoy}
In this subsection we address the question: can CUSTARD \cite{IBMDP} find the optimal policy in $\Pi_{DT}$ with respect to the IBMDP objective of Section \ref{sec:partial-obj}. To that end, we design toy experiments that are amenable to an analysis thanks to their very small size.

\begin{figure}
    \centering
    \subfloat[\label{fig:binary_dt}]{%
      \includegraphics[width=0.45\textwidth]{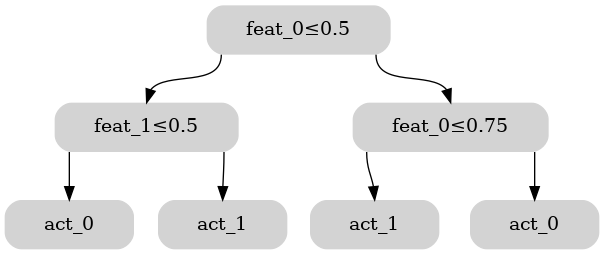}
    }
    \hfill
    \subfloat[\label{fig:zeta_opt}]{%
      \includegraphics[width=0.45\textwidth]{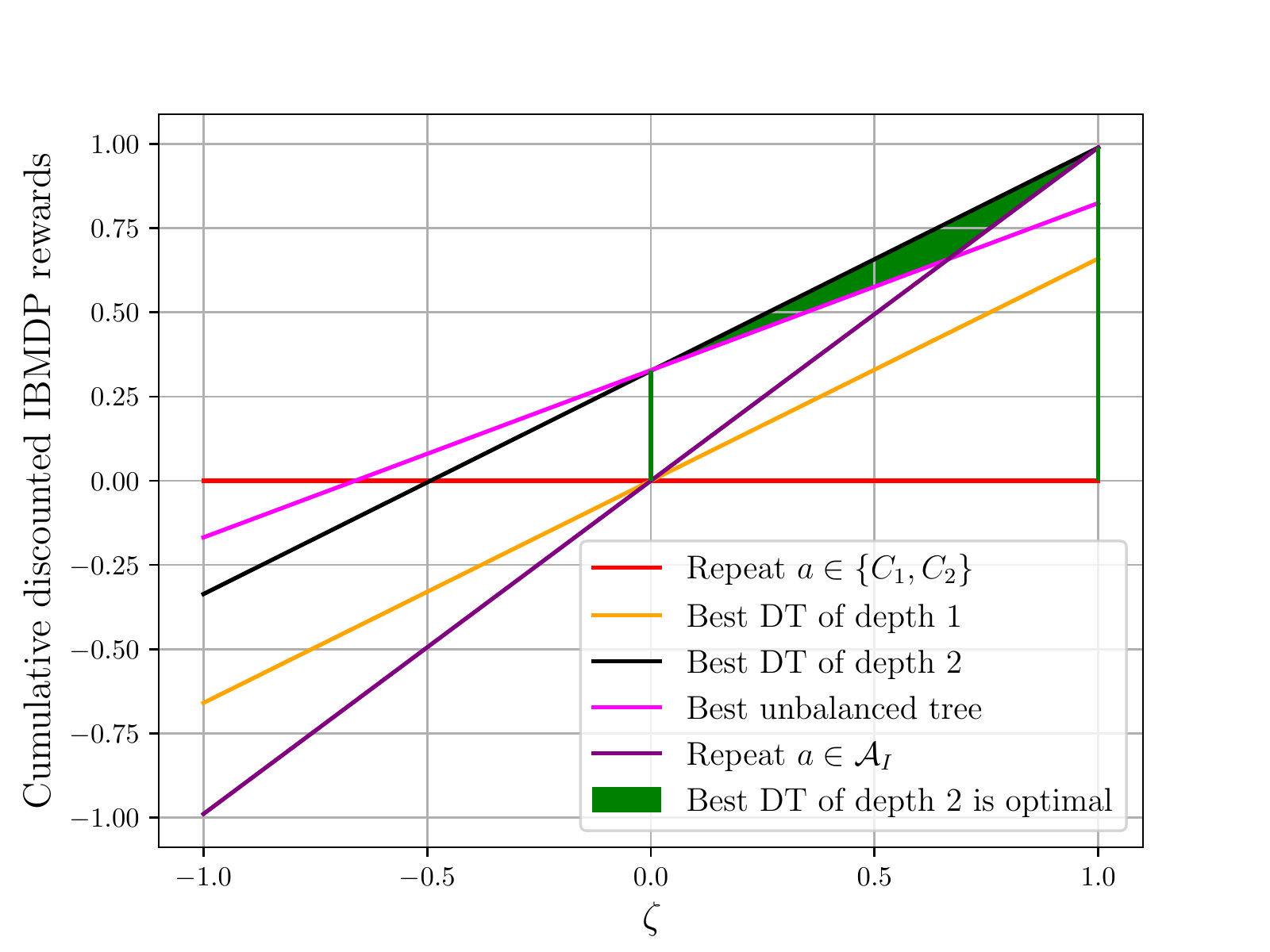}
    }
    \caption{A depth 2 binary DT that is optimal w.r.t to the IBMDP objective when $\zeta = 0.5$ (\ref{fig:binary_dt}) and graphs of the IBMDP reward of different DTs as a function of $\zeta$ (\ref{fig:zeta_opt}).}
    \label{fig:dt-ibmdp}
  \end{figure}

The base tasks are binary supervised classification tasks with 16 different data points and two numerical features in $[0, 1]$. Each data can be perfectly classified using a depth-2 balanced binary tree (see for example Figure \ref{fig:binary_dt}). We generate 5 such classification tasks (hence, we will benchmark CUSTARD to retrieve 5 different DTs). Choosing $\zeta = 0.5$, $\gamma = 0.99$ and $p=1$, induces 5 IBMDPs for which balanced binary DTs of depth 2 are optimal (see Figure \ref{fig:zeta_opt}).

\subsection{CUSTARD to retrieve IBMDP-optimal DTs}

To benchmark CUSTARD, we use \texttt{stable-baselines3} \cite{stable-baselines3} implementations of PPO and DQN and modify them following the definitions of asymmetric $Q$-learning and asymmetric actor-critic (see Section \ref{cust-dqn}). The actor network in PPO is modified to only take feature bounds as inputs while the critic network uses the full state. An additional $Q$-function depending on the full IBMDP state is learned and used as the target network. We use 5 independent runs for each of the 5 different IBDMPs and normalize returns on each IBMDP so that results can be aggregated. Fig.~\ref{fig:custard_bench} shows that none of the agents were able to consistently retrieve the best DT despite the extreme simplicity of the task.

\subsection{Deriving an exact version of CUSTARD}
\label{sec:understanding}

To better understand how theoretically sound asymmetric actor-critic algorithms \cite{Baisero} like CUSTARD PPO fail to retrieve optimal DTs for simple supervised classification tasks, we start from an exact version of CUSTARD where $Q^{\pi}$ and the policy gradient are computed exactly, which is possible in the tabular setup presented next. Note that there do not exist theoretical guarantees for CUSTARD DQN \cite[Section 4.4.2]{baisero22a} equivalent to the one for CUSTARD PPO which is why we focus on the latter.

We introduce a variant of an IBMDP that enforces a maximum depth of the resulting DTs---and ensures that the DT extraction algorithm always terminates. Let this maximum depth be $M + 1$. $M$ is the maximum number of consecutive time-steps during which a policy can select an IGA. We implement this by forcing the policy to take a base action each time it has performed $M$ consecutive IGAs. Interestingly, if $p+1$ is prime (where $p$ is the parameter controlling splitting thresholds in IBMDPs), the state space already provides such information to the policy:
\begin{proposition}
\label{prop:prime}
For an IBMDP, if $p+1$ is prime then there is a mapping $\Omega \mapsto \mathbb{N}$ that maps any feature bound to the number of consecutive IGAs taken since the last base action.
\end{proposition}
In other words, the number of consecutive IGAs since the last base action is directly encoded in the feature bounds~(please see Appendix for proofs of this and all future statements). Thus we can benchmark, on the same IBMDP, algorithms that enforce a maximum tree depth and algorithms such as CUSTARD \cite{IBMDP} that do not.

Having fixed a maximum tree depth $M + 1$, the number of unique feature bounds, i.e.\@ the cardinality of the observation space $|\Omega|$, becomes finite and is at most $(2pd)^M$. Here $pd = |\A_I|$ is the number of IGAs available at any time~(if available at all) and the factor of $2$ stems from the two possible state transitions following an IGA. Since the state-action space of an IBMDP becomes finite, and its transition and reward functions are known, one can compute the policy gradient exactly. This will let us investigate whether the sub-optimal performance of CUSTARD is due to approximation errors---e.g. introduced by the learned value function---or if it is a limitation of the gradient descent approach in itself. 

Because $\Omega$ is finite, we can additionally implement policy gradient on tabular policies which would eliminate any representation error of the policy. With a slight abuse of notation, we let in this case $\theta(o,a)$ be the logit of observation-action pair $(o, a)$, i.e. $\pi(a|o)\propto \exp(\theta(o,a))$. By a straightforward application of the chain rule on Lemma~C.1 of \cite{Agarwal19} we obtain:
\begin{proposition} 
\label{prop:grad}
Let $\theta \in \mathbb{R}^{\Omega\times\A'}$ be the logits of a tabular reactive policy of the IBMDP, then:
\begin{align}
\label{eq:polgradtab}
    \frac{\partial J(\pi_\theta)}{\partial \theta(o, a)}
    &={\sum_{s\in \SC'}} {1}_{O(s) = o}\frac{p^{\pi_\theta}(s)}{1-\gamma}\pi_\theta(a|o) A^{\pi_{\theta}}(s, a).
\end{align}
\end{proposition}
Here ${1}_{O(s) = o} = 1$ if the feature bound part of $s$ is $o$, 0 otherwise.
\subsection{Ablation study}\label{sec:abla}
\begin{figure}
\centering
    \subfloat[\label{fig:custard_bench}]{%
      \includegraphics[width=0.45\textwidth]{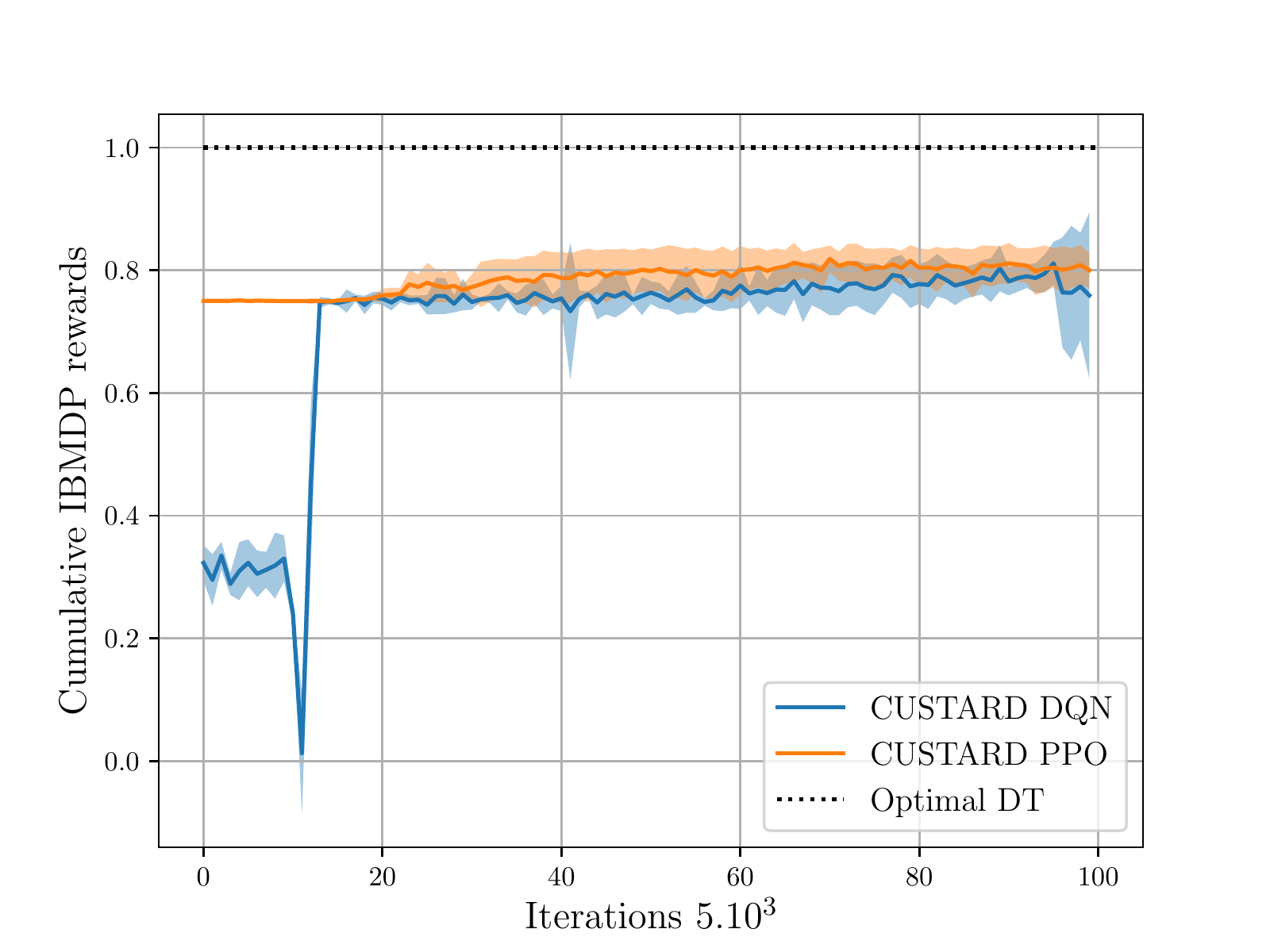}
    }
    \subfloat[\label{fig:ablation}]{
      \includegraphics[width=0.45\textwidth]{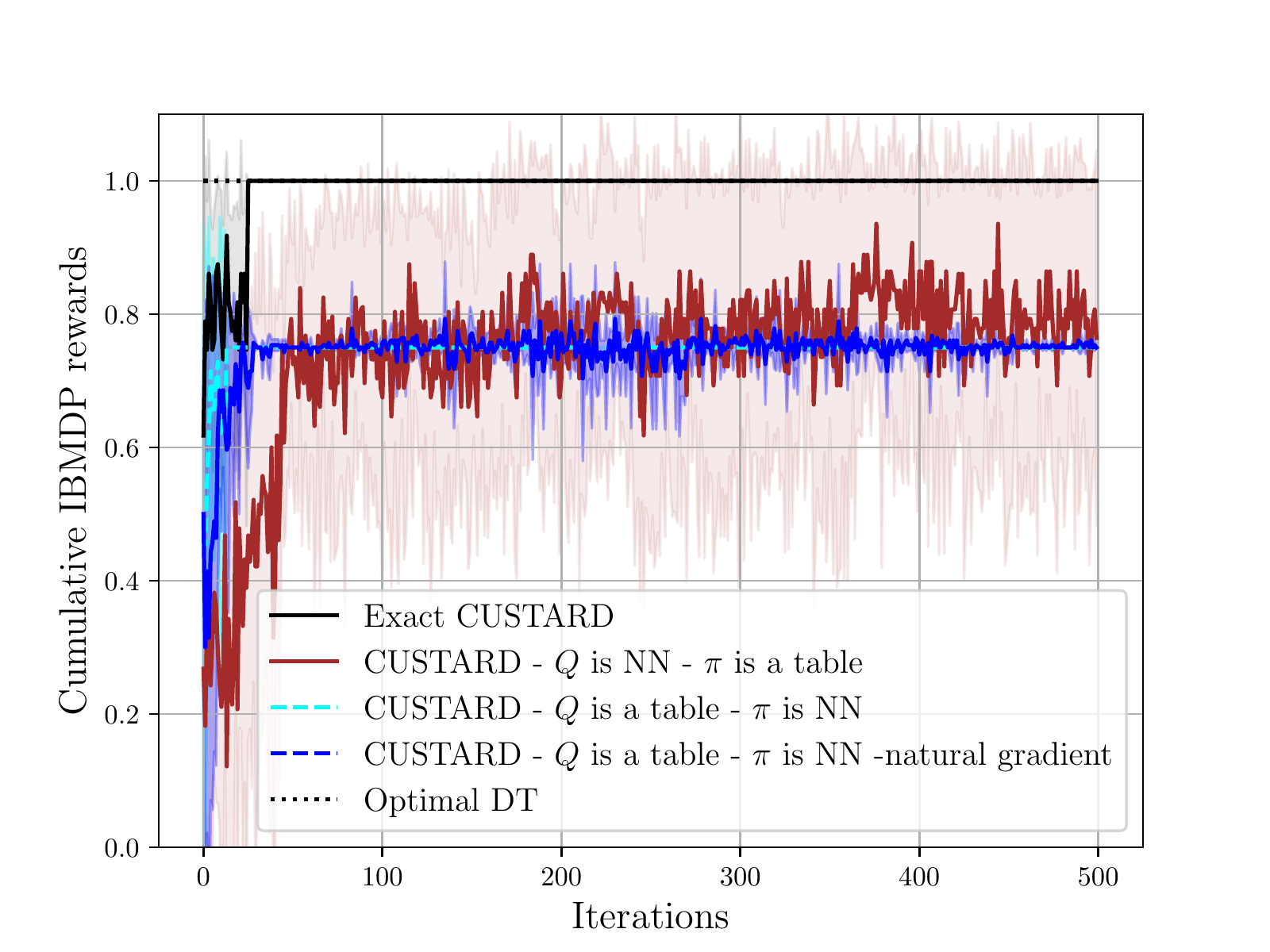}
    }
    \caption{Study of CUSTARD algorithms ability to retrieve DTs for simple supervised classification tasks by solving IBMDPs. In Figure \ref{fig:custard_bench}, we plot the IBMDP cumulative reward during training of CUSTARD as well as the IBMDP cumulative reward of the IBMDP-optimal DT. On Figure \ref{fig:ablation}, we plot the IBMDP cumulative reward during training of an exact version of CUSTARD, the IBMDP cumulative reward of the IBMDP-optimal DT, as well as the cumulative reward of different approximated versions of the exact CUSTARD.}
    \label{fig:aac}
\end{figure}
Starting from the exact CUSTARD algorithm defined above, we perform an ablation study to get to a CUSTARD algorithm similar to \cite{IBMDP}. Algorithms are tested on the same IBMDPs as in Section \ref{sec:xptoy}. The main features ablated from the exact CUSTARD are: \\
\textbf{Using an approximated $\hat{Q}^\pi$-function} instead of a $Q^\pi$-table updated exactly. In that case, $\hat{Q}^\pi$ is a neural network similar to the one in CUSTARD PPO. \\ \textbf{Using neural network for the policy $\pi$} instead of a table. In that case, the policy network is similar to the one in CUSTARD PPO. 

The results of the ablation are presented on Figure \ref{fig:ablation}. The exact CUSTARD algorithm consistently finds the optimal policy. This is important as it means that the partially observable framework of CUSTARD is not the reason for the poor results, at least when implemented exactly. In practice however, we find that both the approximation errors of the neural Q-function and the policy representation error hinder performance. Indeed, when the policy is encoded by a neural network in place of a table, the aggregated cumulative IBMDP reward converges to sub-optimal values after just a few iterations. When the $Q^\pi$ function is encoded by a neural network, some instances of the associated CUSTARD algorithm converged to the optimal policy---reflected by the higher standard deviations on Fig. \ref{fig:ablation}, but many did not.

\section{IBMDP-optimal policies by solving fully observable MDPs}

The main result of our work is to show that when using the IBMDP framework to learn a DT for a supervised classification task, there is no need to use partially obesrvable RL and that it is sufficient to use classical RL. We first define a new MDP and then show that a policy maximizing the expected cumulative reward of this MDP also maximizes the IBMDP reward. 
\subsection{Observation-IBMDP}
\label{sec:oibmdp}
Let us consider a base Classification MDP $\langle \mathcal{X}, \{C_1, ..., C_K\}, R, T, \gamma \rangle$. 
and an associated IBMDP $\langle \SC^{\prime}, \A^{\prime}, R^{\prime}, T^{\prime}, \zeta, p, \gamma\rangle$. An Observation-IBMDP (OIBMDP) $\langle \Omega, \A^{\prime}, R^{\prime\prime}, T^{\prime\prime}, \zeta, p, \gamma\rangle$ is defined as follows:
\paragraph{State space}
\vskip -0.1in
The state space is the space of possible feature bounds $\Omega \subsetneq [0,1]^{2d}$.
\paragraph{Action space} 
\vskip -0.1in
The action space is $\A_I$, the same as in the given IBMDP.
\paragraph{Reward function}
\vskip -0.1in
Assume the current state of the MDP is $o = (L_1, ..., L_d, U_1, ..., U_d)$.
\begin{itemize}
    \item $a \in \A_I$ : The reward for taking an IGA is still $\zeta$.
    \item $a = C_h \in \{C_1, ..., C_K\}$:  We denote $\mathcal{X}^{C_h}_{o}$ the set of all $x_i$ such that $y_i = C_h$ and $ L_k \leq x_{ik} \leq U_k$ for all $k$. Similarly,   We denote $\mathcal{X}^{\bar{C_h}}_{o}$ the set of all $x_i$ such that $y_i \neq C_h$ and $ L_k \leq x_{ik} \leq U_k$ for all $k$. So $R^{\prime\prime}(o, C_h) = \frac{|\mathcal{X}^{C_h}_{o}|- |\mathcal{X}^{\bar{C_h}}_{o}|}{|\mathcal{X}^{C_h}_{o}| +|\mathcal{X}^{\bar{C_h}}_{o}|}$
\end{itemize}
\paragraph{Transition function}
Assume the current state of is $o = (L_1, ..., L_d, U_1, ..., U_d)$.
\begin{itemize}
    \item $a = C_h \in \{C_1, ..., C_K\}$: $T(o, C_h, (0,..., 0, 1, ...,1)) = 1$
    \item $a = (k, \frac{u}{p+1})\in \A_I$: We denote $v = \frac{u}{p+1}(U_k - L_k) + L_k$. The MDP will transit to $o_{inf} = (L_1, ..., v, ..., L_d, U_1, ..., U_d)$  (resp. $o_{sup} = (L_1, ..., L_d, U_1, ..., v, ..., U_d)$) with probability $\frac{|\mathcal{X}_{o_{inf}}|}{|\mathcal{X}_{o_{inf}}| + |\mathcal{X}_{o_{sup}}|}$ (resp. $\frac{|\mathcal{X}_{o_{sup}}|}{|\mathcal{X}_{o_{inf}}| + |\mathcal{X}_{o_{sup}}|}$)
\end{itemize}
\begin{theorem}
\label{theo:obs}
    Any optimal policy of the OIBMDP has the same policy return as $J(\pi^*)$ in the IBMDP. As such, any policy optimal w.r.t the OIBMDP reward is optimal w.r.t a certain interpretability-performance trade-off.
\end{theorem}

\subsection{Avoiding to learn large trees}
We will observe from experimental results that CUSTARD \cite{IBMDP} tends to learn large trees (more than 10 decision nodes). To avoid this problem, we learn an OIBMDPs with a maximum tree depth $M_{max}$ and with $p$ a prime number to leverage Proposition \ref{prop:prime}. Thus given the current observation $o = (L_1, ..., L_d, U_1, ..., U_d)$, the current depth is $M = \sum_{i = 1}^d \log_2(\frac{1}{U_i - L_i})$. When an agent learning in an OIBMDP observes $o$ such that $M \geq M_{max}$, it takes an action in $\{C_1, ..., C_K\}$ equivalent to adding a leaf node to the learned DT. 

\subsection{DQN Decision Tree}\label{dqn-dt}
\begin{figure}[h!]
    \centering
    \includegraphics[scale=0.4]{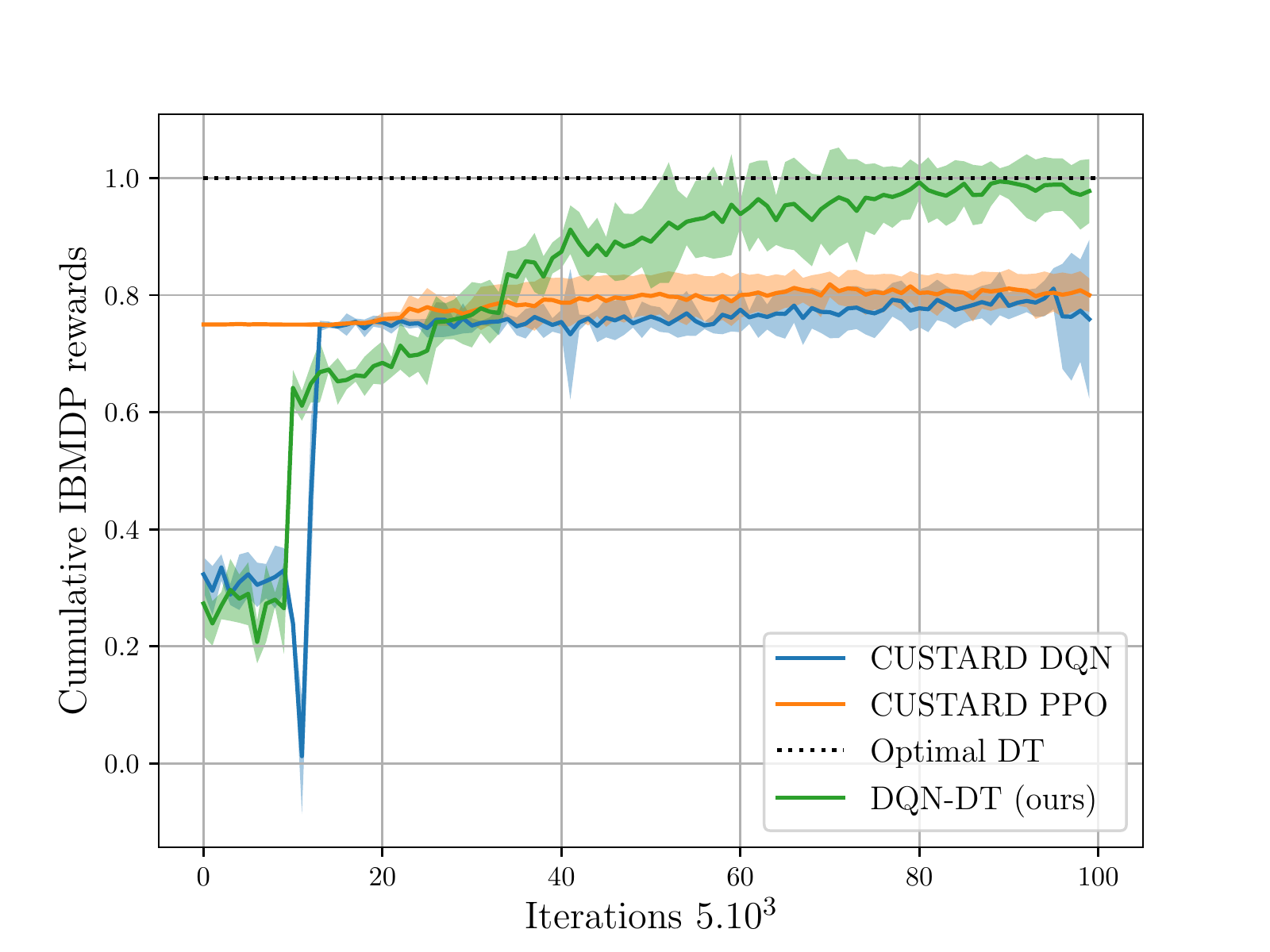}
    \caption{Comparison of CUSTARD and DQNDT on the simple binary classification benchmark of Section \ref{sec:xptoy}.}
    \label{fig:bench_dqn_dt}
\end{figure}

We now present a DQN \cite{mnih2015human} variant to learn DTs for OIBMDPs with a given max depth. As guaranteed by Theorem \ref{theo:obs}, any optimal policy of the OIBMDP is equivalent to the DT with the optimal interpretability-performance trade-off encoded by an IBMDP reward function. However existing algorithms returning optimal policies for MDPs \cite{sutton2018reinforcement} require the full state space to be stored in memory and the state space of OIBMDPs is of size $(2pd)^{M_{max}}$ with elements in $\mathbb{R}^d$. Experimentally, we used Policy Iteration \cite{sutton2018reinforcement} to consistently retrieve the optimal DT for the benchmarks of Section \ref{sec:xptoy}. However, when the number of features $d$ and $M_{max}$ grow, Policy Iteration becomes intractable. We have also tried Q-learning and SARSA \cite{sutton2018reinforcement} but did not manage to solve the benchmark of Section \ref{sec:xptoy}. To overcome this challenge we go for the Deep RL algorithm DQN \cite{mnih2015human} that we modify so that $\pi(o) = \underset{a \in \{C_1, ..., C_K\}}{\operatorname{argmax}} Q(o, a)$ when $M \geq M_{max}$. It is clear from Figure \ref{fig:bench_dqn_dt} that DQNDT outperforms CUSTARD on the simple benchmark of Section \ref{sec:xptoy}
\section{Experiments on Supervised Classification Datasets}
In this section, we apply DQNDT on UCI \cite{UCI} datasets. We compare the learned DTs of DQNDT with trees learned by CUSTARD \cite{IBMDP} with respect to the interpretability-performance trade-off.  
\begin{table}
\caption{UCI datasets: number of training samples, number of features per sample, and number of classes.}\label{tab:uci}
\vskip 0.2in
\centering
\begin{tabular}{|l|l|l|l|}
\hline
& $|\mathcal{X}|$ & $d$ & $K$ \\ \hline
\texttt{Wine}     &178   &13 &3  \\ \hline
\texttt{Diabete}  &520  &16 &2   \\ \hline
\texttt{Banknote} &1372   &4   &2   \\ \hline
\end{tabular}
\end{table}
\subsection{Reproducibility statement}
All the code to reproduce the experiments is given in the \underline{anonymous github} (footnote \ref{git}). All experiments were run multiple times on independent seeds. All the versions of the necessary \texttt{python} libraries are given in a \texttt{requirements.txt} file. 

All implementations of CUSTARD and DQNDT are available in the \underline{anonymous} \underline{github} (footnote \ref{git}). We modify the source code of \texttt{stable-baselines3} \cite{stable-baselines3} implementations of PPO and DQN with the modification of Sections \ref{cust-dqn} and \ref{dqn-dt}. All hyperparameters of CUSTARD and DQNDT are the default hyperparameters of PPO and DQN from \texttt{stable-baselines3}. 
\subsection{Experimental setup}
For each UCI dataset, we try 6 different values for the cost of building decision nodes: $\zeta \in \{-1, -0.6, -0.2, 0.2, 0.6, 1\}$. The splitting parameter $p$ is always 1. Hence  we solve 6 different IBMDPs (resp. OIBMDPs) with CUSTARD (resp. DQNDT). Each DT learning agent is run 5 times with 2 million timesteps. For each UCI dataset and each $\zeta$, we analyse the best DTs obtained with each agent. We report the accuracy and the number of nodes of the best DTs and plot the interpretability-performance trade-off. For DQNDT, we fix the maximum tree depth $M_{max}$ to 4, 5, and 5 for \texttt{Wine, Diabete, Banknote} respectively. We also compute DTs using CART (we use the implementation of \cite{scikit-learn}, and fix the \texttt{max\_depth} parameter to 4 or 5).
\subsection{Interpretability-Performance Trade-Offs}
\begin{figure}
\centering
    \subfloat[\texttt{Wine Trade-off}]{%
      \includegraphics[width=0.5\textwidth]{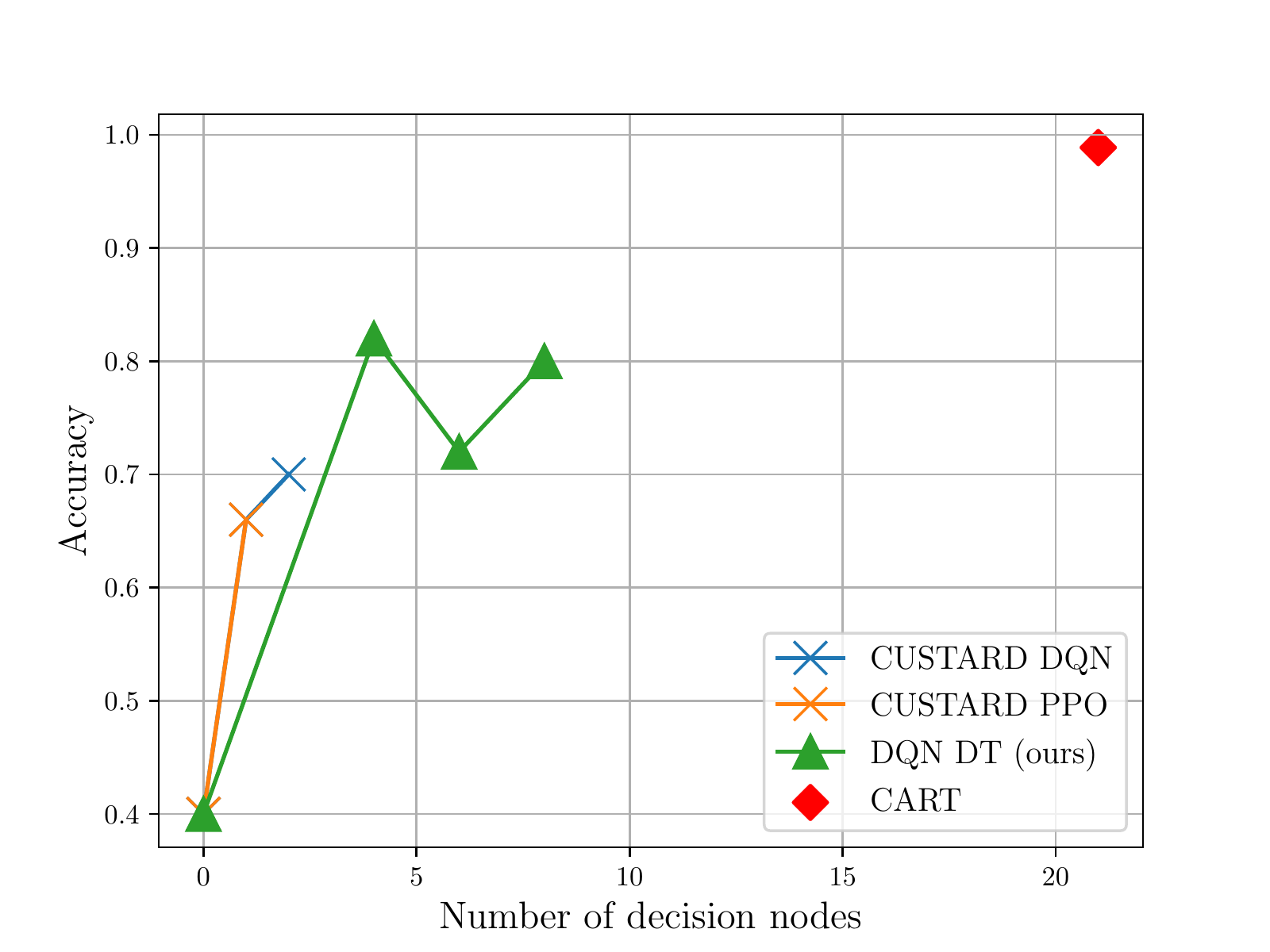}
    }
    \subfloat[\texttt{Wine DT: accuracy 0.82, decision nodes 4}]{%
      \includegraphics[width=0.4\textwidth]{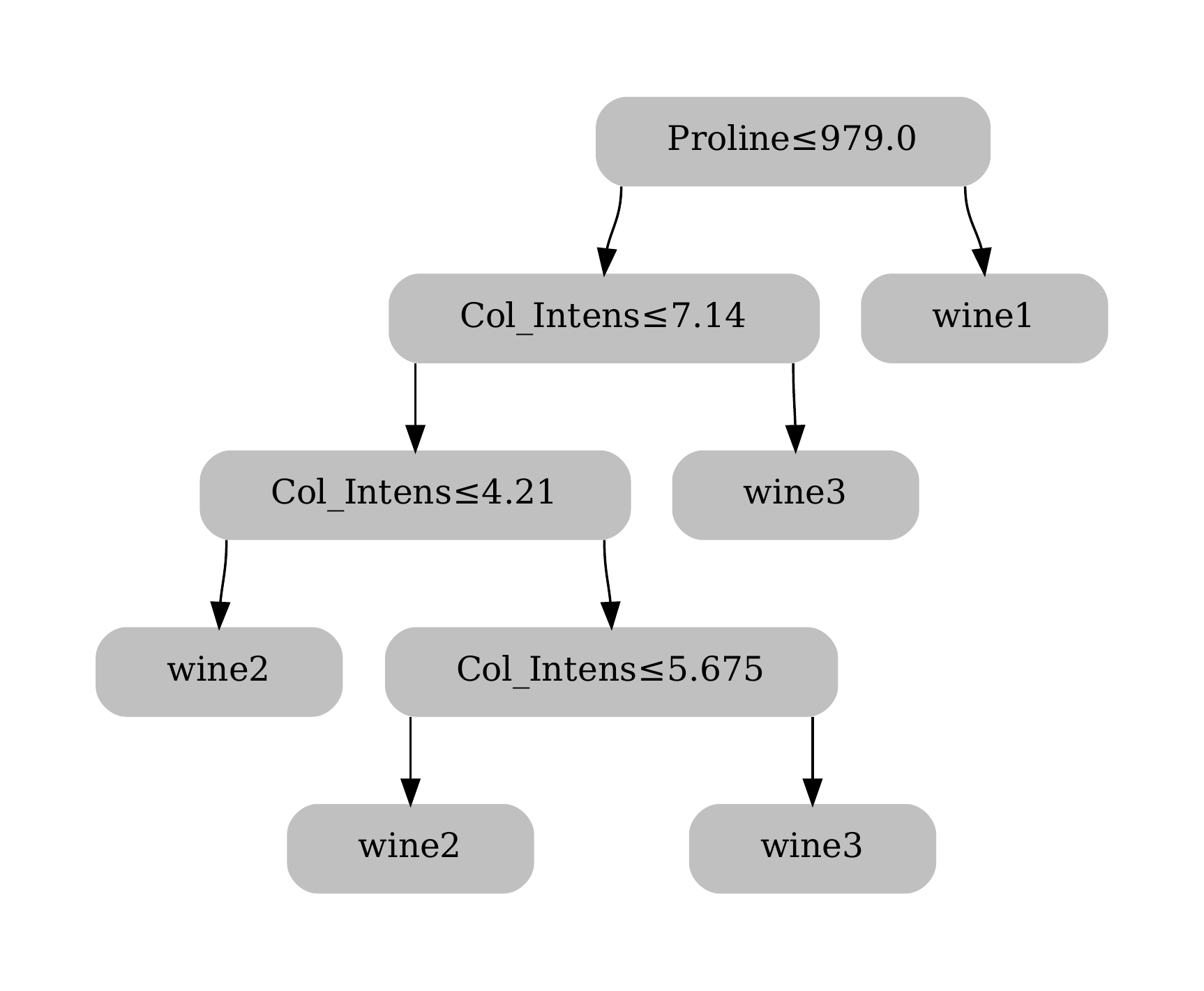}
    }\\
    \vskip -0.05in
    \subfloat[\texttt{Banknote Trade-off}]{
      \includegraphics[width=0.5\textwidth]{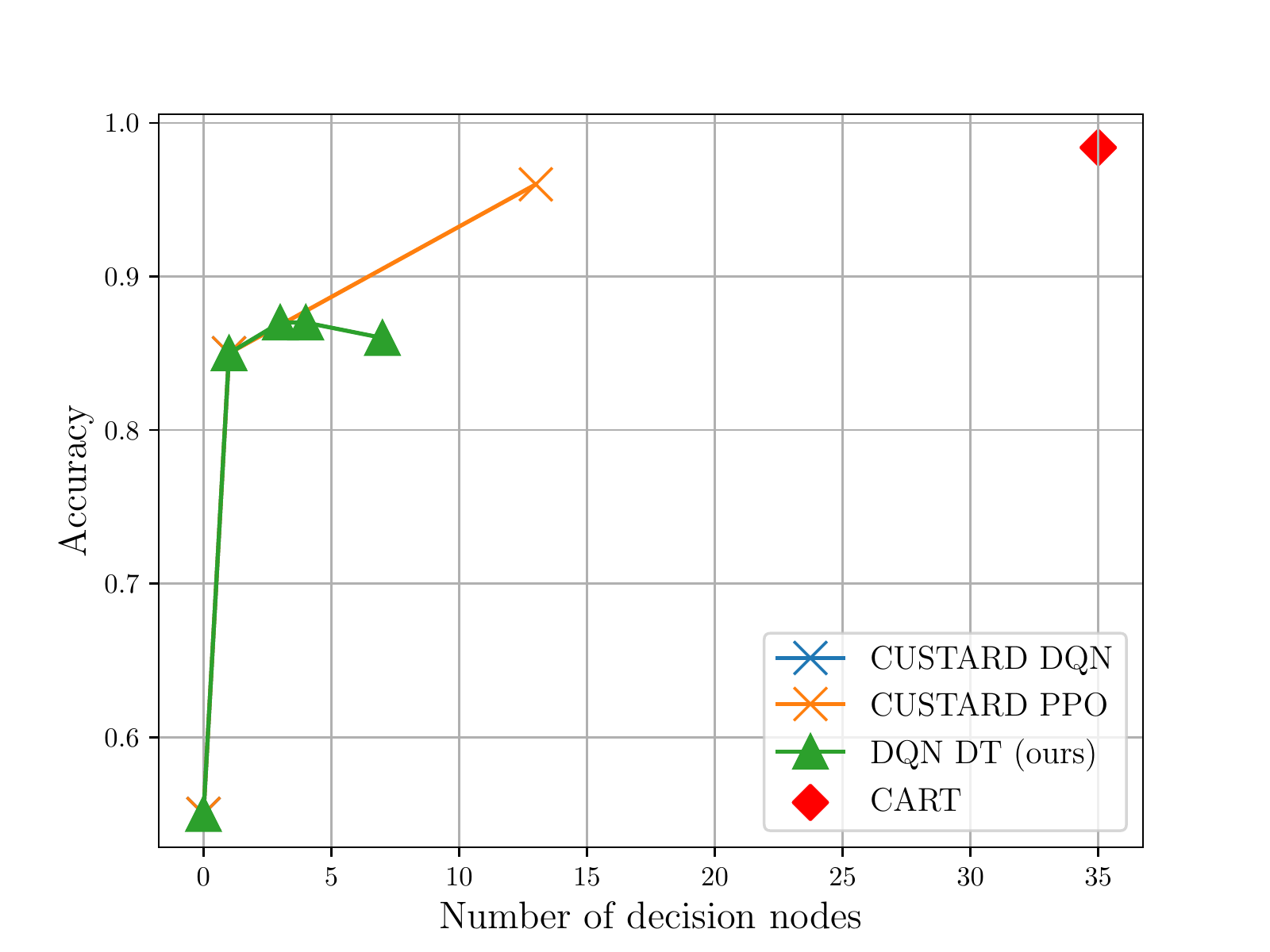}
    }
    \subfloat[\texttt{Banknote DT: accuracy 0.87, decision nodes 4}]{
      \includegraphics[width=0.4\textwidth]{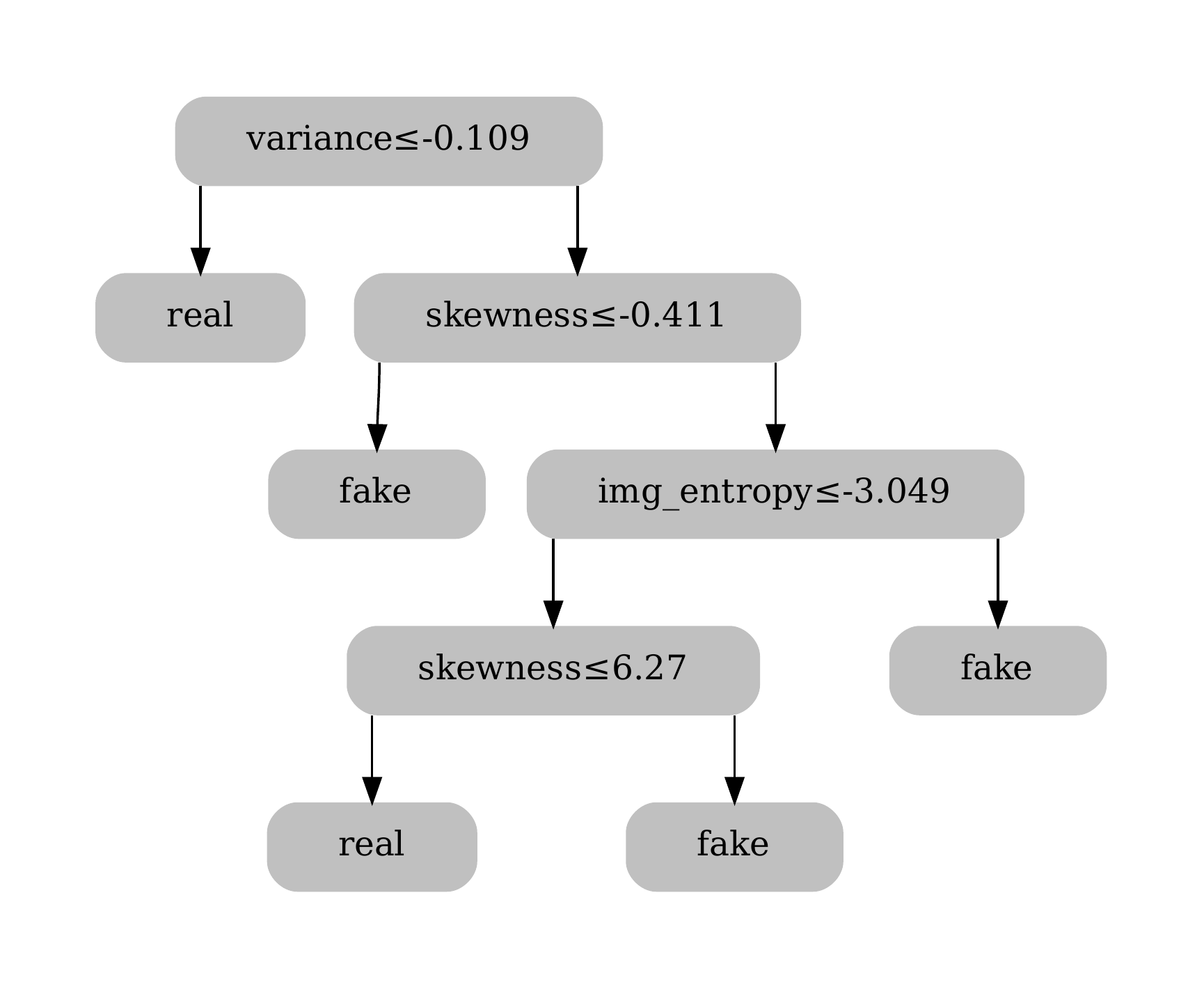}
    }\\
    \vskip -0.05in
    \subfloat[\texttt{Diabete Trade-off}]{
      \includegraphics[width=0.5\textwidth]{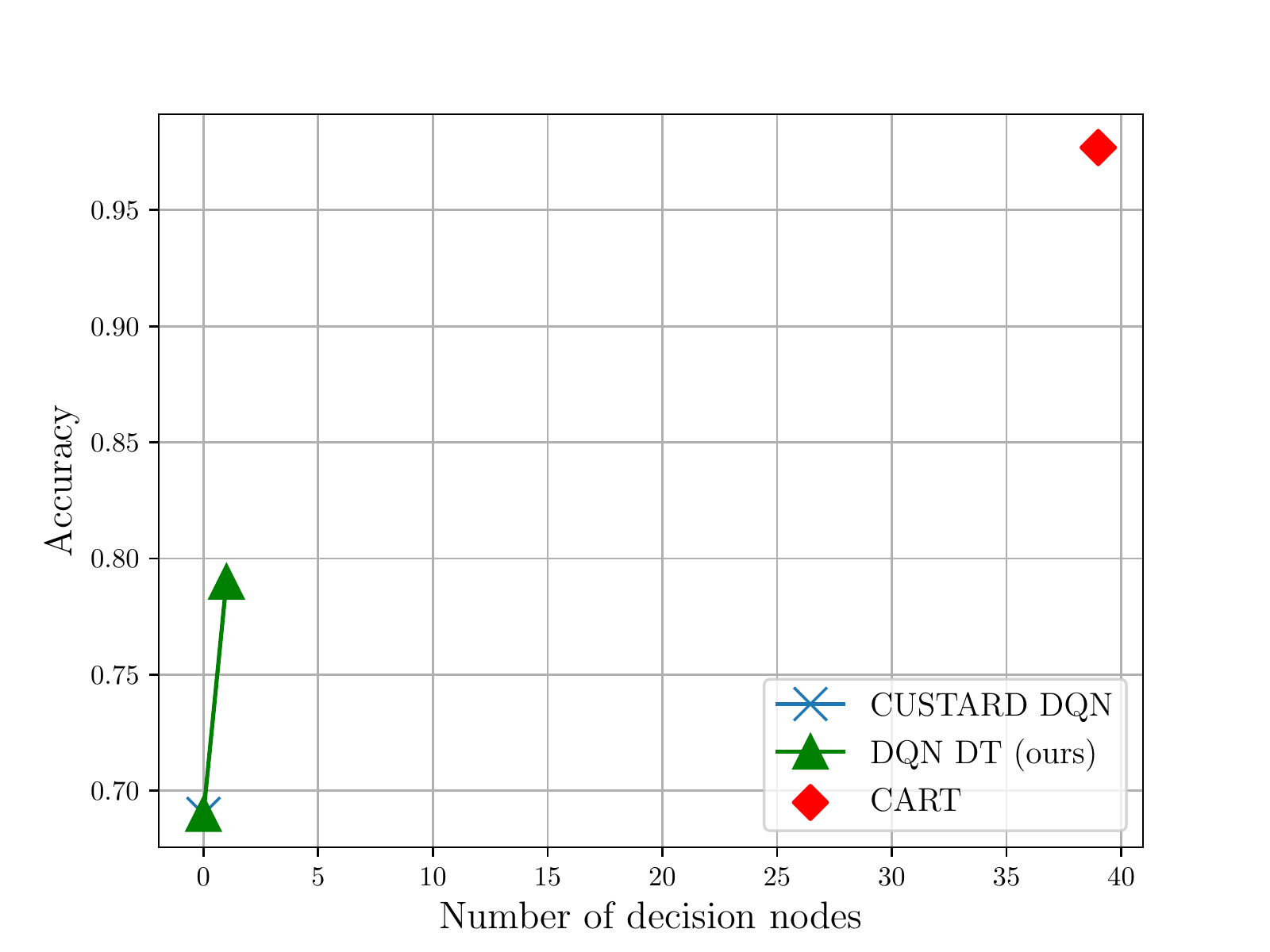}
    }
    \subfloat[\texttt{Diabete DT: accuracy 0.8, decision nodes 1}]{
      \includegraphics[width=0.4\textwidth]{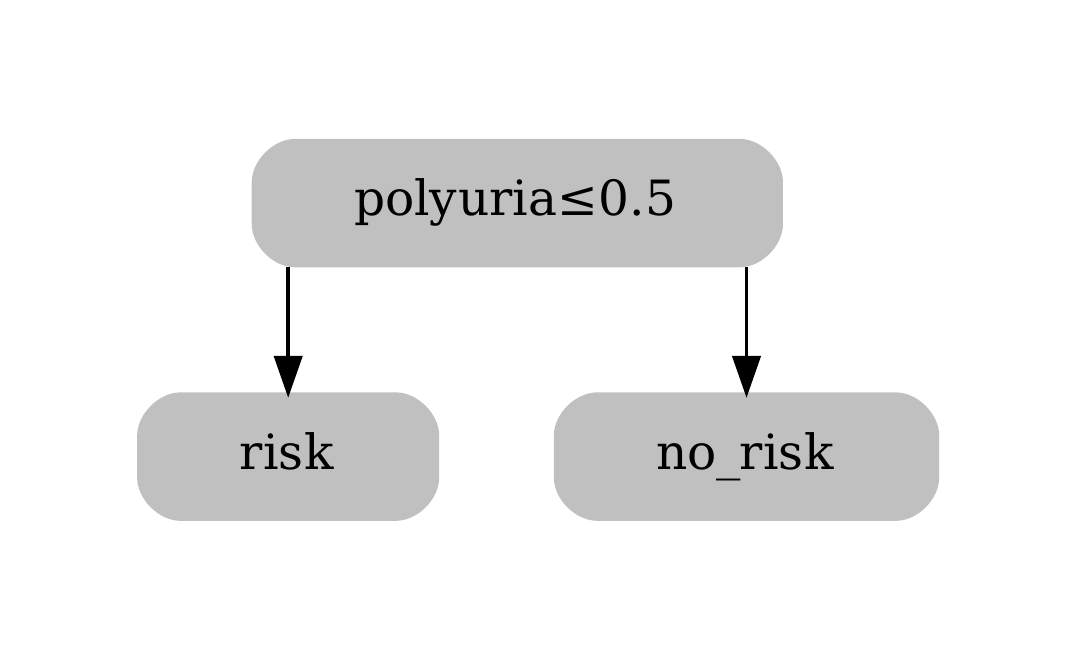}
    }\\
    \caption{Interpretability-performance trade-offs of DTs learned using CUSTARD, DQNDT, or CART; DTs learned with DQNDT.}
    \label{fig:tradeoffs}
\end{figure}
From Figure \ref{fig:tradeoffs}, it is clear that CUSTARD learns either very small trees (2 decision nodes at most for the \texttt{Wine} dataset) or very large trees (up to 13 decision nodes for the \texttt{Banknote} dataset). And in general, for $\zeta \geq 0.6$, CUSTARD learns to repeat IGAs indefinitely. DQNDT is able to return DTs with a wider range of interpretability-performance trade-offs (DTs with 2, 4, 6, 7, 8 decision nodes). The trees returned by DQNDT are always limited in depth. As expected, CART follows a greedy approach and simply maximises the accuracy of DT resulting in non-interpretable trees with many decision nodes. All computed trees are available in the \underline{anonymous github} (footnote \ref{git}).

\section{Future work}

\subsection{Beyond DTs} Scaling to larger IGA spaces would allow for finer splittings at decision nodes, but could also ease the generalization of IBMDPs to more general classes of interpretable models. At the very least one could consider test nodes that also test if a feature value is within closed intervals $[v_1, v_2]$ or for example could consider combinations of more than one feature. As long as one can write the associated transition function, it is very likely that our results would extend to such settings too. 

\subsection{Beyond supervised learning}
Finally, we stress out that provably convergent algorithms for finding an optimal IBMDP policy when the base task is a sequential decision making problem remains an open question. Perhaps one intermediate step is to study the learning of DT policies by imitation learning as in \cite{Viper}, which can be reduced to a sequence of supervised learning problems.

\section{Conclusion}

In this work we analysed and evaluated the recently proposed IBMDP framework \cite{IBMDP} that tackles the interesting problem of learning compact DTs with RL. We showed that CUSTARD \cite{IBMDP} can be seen as asymmetric RL \cite{Pinto,Baisero,baisero22a}. We showed experimentally that CUSTARD fails to retrieve the optimal DT for IBMDPs of simple supervised classification base tasks. One of the main contribution of the paper is to show that this problem can be reformulated into a fully observable MDP. To scale to supervised classification tasks with large feature dimension and to avoid the learning of arbitrarily large DTs, we proposed a variant of DQN \cite{mnih2015human}: DQNDT. We showed experimentally that DQNDT is able to learn DTs with more interpretability-performances trade-offs than CUSTARD on UCI \cite{UCI} datasets. Our work opens up a large avenue for future research to go beyond greedy search algorithms and explore the full space of DTs and similar models of discrete nature. 

\section{Ethical statement}

The ethical implications of this work are the same as the DQN algorithm \cite{mnih2015human}. Furthermore since DQNDT learns an interpretable DT, we believe our work offers a new way to solve classification tasks in an ethical manner.
\section{Acknowledgements}
Hector Kohler acknowledges the funding from ANR AI\_PhD@Lille for his PhD. Philippe Preux acknowledges the support of the M\'etropole Europ\'eenne de Lille (MEL), ANR, Inria, Universit\'e de Lille, through the AI chair Apprenf number R-PILOTE-19-004-APPRENF. We also acknowledge the outstanding working environment provided by Inria in the Scool research group \footnote{https://team.inria.fr/scool/}.

\newpage

\appendix
\section{Proofs}
\label{app:proofs}
\subsection{Proposition~\ref{prop:prime}: tree depth information in feature bounds}
\label{app:prime}
We want to show that in an IBMDP, if $p+1$ is prime, then the number of information-gathering actions performed since the last base action is encoded in the feature bounds part of the state. Without loss of generality we only study the case of a single feature, showing that $l = U - L$, the difference between the upper and lower bound of the feature, fully determines the number of information-gathering actions taken since the last base action. The extension to multiple features is trivial by addition of each feature's inferred number of information-gathering actions. 

Let $n$ be the number of information-gathering actions taken since the last base action. Let $l_n$ be the difference between upper and lower bounds of the feature after these $n$ information-gathering actions. Clearly $n = 0 \Leftrightarrow l_n = 1$. When $n > 0$, $l_n$ is given by $l_n = \overset{n}{\underset{k=1}{\prod}}\frac{h_k}{p+1}$, where $h_k\in \{1,...,p\}$ for each $k$. 
We want to show that if $p+1 \text{ is prime} \implies \nexists (i,j): i < j \text{ and } l_i = l_j$.
\begin{align*}
    \text{Suppose } &\exists (i,j): 0<i<j \text{ and } l_i=l_j,\\
    & \implies \frac{\overset{i}{\underset{k=1}{\prod}}h_k}{(p+1)^{i}} = \frac{\overset{j}{\underset{k=1}{\prod}} h'_k}{(p+1)^{j}},\\
    & \implies \overset{i}{\underset{k=1}{\prod}}h_k = (p+1)^{j - i} \overset{j}{\underset{k=1}{\prod}} h'_k,  \\
    & \implies \overset{i}{\underset{k=1}{\prod}}h_k = (p+1) (p+1)^{j - i - 1} \overset{j}{\underset{k=1}{\prod}} h'_k,\\
\end{align*}
But that is impossible since the left-hand side is the product of non-zero natural numbers $< p + 1$ and the right-hand side is the product of non-zero natural numbers containing the prime number $p+1$. $\square$

\subsection{Proposition~\ref{prop:grad}: gradient of soft-max reactive policy}
\label{app:gradtab} 
For an MDP $\langle \SC, \A, R, T, \gamma \rangle$ with finite state-action spaces, Lemma~C.1 of \cite{Agarwal19} showed that for tabular soft-max policies
\begin{align}
    \label{eq:tabmdp}
    \frac{\partial J(\pi_\theta)}{\partial \theta(s, a)}
    &=\frac{1}{1-\gamma}p^{\pi_\theta}(s)\pi_\theta(a|s) A^{\pi_{\theta}}(s, a),
\end{align}
where $\theta(s, a)$ is the logit parameter of the policy. Now extending this MDP into an IBMDP $\langle \SC^{\prime}, \A^{\prime}, R^{\prime}, T^{\prime}, \zeta, p, \gamma\rangle$ with a fixed maximum depth as defined in Sec.~\ref{sec:understanding}, the state space remains finite. Let $s = (\phi, o) \in \SC'$ be a state of the IBMDP, with $o \in \Omega$ where $\Omega$ is the finite set of reachable feature bounds. We are interested in tabular policies parameterized by logits $\theta' \in  \mathbb{R}^{\Omega \times \A'}$ such that $\pi_{\theta'}(a | s) \propto \exp(\theta'(o, a))$. That is, the main difference with the setting of Eq.~\eqref{eq:tabmdp} is that a given logit $\theta'(o, a)$ is shared between several states and provides the unormalized log-probability of taking action $a$ in all states $s \in \SC'$ such that their feature bounds is $o$, i.e. such that $O(s) = o$. Informally, we can decompose this map $\theta' \mapsto J(\pi_{\theta'})$ going from logits in $\mathbb{R}^{\Omega \times \A'}$ to a policy return into the composition $\theta' \mapsto \theta \mapsto J(\pi_\theta)$, where the first map maps logits in $\mathbb{R}^{\Omega \times \A'}$ into logits in $\mathbb{R}^{\SC' \times \A'}$ according to $\theta(s, a) = \theta'(O(s), a)$. By the chain rule, we have $\nabla_{\theta'}J(\pi_{\theta'}) = H(\theta, \theta')^T\nabla_{\theta}J(\pi_\theta)$ where $H$ is the Jacobian of the map $\theta'\mapsto \theta$. This Jacobian will have a value of 1 at row $(s, a)$ and column $(O(s), a)$ for all $s$ and $a$ and is 0 otherwise. Thus the product simply becomes
\begin{align}
\label{eq:polgradtab_int}
    \frac{\partial J(\pi_\theta)}{\partial \theta'(o, a)}
    &={\sum_{s\in \SC'}} {1}_{O(s)=o}\frac{\partial J(\pi_\theta)}{\partial \theta(s, a)}\bigg|_{\theta(s, a) = \theta'(o, a)}
\end{align}

Combining Eq.~\eqref{eq:tabmdp} and Eq.~\eqref{eq:polgradtab_int} completes the proof of Eq.~\eqref{eq:polgradtab}.
\subsection{Theorem~\ref{theo:obs}: problem equivalence with the OIBMDP}
\label{ann:equi}
A stochastic reactive policy of an IBMDP $\pi: \Omega \mapsto \Delta\A'$, where $\Delta\A'$ is the set of all probability distributions over $\A'$, can also act on the OIBMDP (Sec.~\ref{sec:oibmdp}) since the state and action spaces of the OIBMDP  are respectively $\Omega$ and $\A'$. We will show in this section that any such policy has the same policy return in the IBMDP and the OIBMDP. Indeed, the construction of the OIBMDP's transition function is such that feature bounds are visited with the same frequency in the IBMDP and the OIBMDP, while the reward at state $o$ of the OIBMDP is the average over the state rewards of the IBMDP that 'fall' within the feature bounds of $o$.

The proof only holds when the base MDP of the IBMDP is a supervised task because we have that for any reactive policy acting in the IBMDP, $Pr(s_t = s|o_t = o) = p(s|o)$. That is, for a policy $\pi$ acting in the IBMDP, if at time-step $t$ the observation part of the state is $o$, then the distribution of the random variable $s_t$ follows the fixed distribution $p(s|o)$. This is trivial to see because on one hand, information gathering actions in an IBMDP only influence the observation part of the state, and on the other hand, all base actions induce the same next state distribution since in the supervised setting, the next data point is sampled randomly from the dataset independently from the last prediction~(base action). Thus one can easily show by induction that for any policy $\pi$, $Pr(s_t = s|o_t = o)$ is independent of $\pi$, since the base state is independent of the actions of the policy.      

Following Sec.~\ref{sec:oibmdp}, one can see that the reward and transition functions of the OIBMDP can be written as 
\[R''(o, a) = \sum_{s\in\SC'}p(s|o)R'(s,a),\] and 
\[T''(o_t, a_t, o_{t+1}) = \sum_{s, s' \in \SC'}{1}_{O(s') = o_{t+1}}p(s|o_t)T'(s, a_t, s'),\]
where $p(s|o) = \frac{1}{|\mathcal{X}_{o}|}$ if $O(s) = o$ and the base state of $s$ is in $\mathcal{X}_{o}$ and is $0$ otherwise.

Let $o_t$ and $v_t$, $t\geq0$ be the feature bounds random variables as $\pi$ acts on the IBMDP and the OIBMDP respectively. We will show that for all $o\in\Omega$ and $t\geq 0, Pr(o_t = o) = Pr(v_t = o)$. By induction, it is true for $t=0$ since the feature bounds are all initialized to $(0,1)$  for both the IBMDP and OIBMDP. Assume it is true for $t$ then, 
\begin{align}
    Pr(o_{t+1} = o') &= \sum_{a\in\A'}\sum_{s, s'\in\SC'}Pr(s_t = s)\pi(a|O(s)) T'(s, a, s') {1}_{O(s') = o'}\\
    &= \sum_{o\in\Omega}\sum_{a\in\A'}\sum_{s, s'\in\SC'}Pr(s_t = s|o_t = o)Pr(o_t=o)\pi(a|O(s)) T'(s, a, s') {1}_{O(s') = o'}\\
    &= \sum_{o\in\Omega}Pr(o_t=o)\sum_{a\in\A'}\pi(a|o)\sum_{s, s'\in\SC'}Pr(s_t = s|o_t = o) T'(s, a, s') {1}_{O(s') = o'}\\
    &= \sum_{o\in\Omega}Pr(o_t=o)\sum_{a\in\A'}\pi(a|o)\sum_{s, s'\in\SC'}p(s|o) T'(s, a, s') {1}_{O(s') = o'}\\
    &= \sum_{o\in\Omega}Pr(v_t=o)\sum_{a\in\A'}\pi(a|o)\sum_{s, s'\in\SC'}T''(o, a, o')\\
    &=Pr(v_{t+1} = o')
\end{align}
Now let $J_O(\pi)$ be the policy return of $\pi$ when acting on the OIBMDP. We have 
\begin{align}
J(\pi) &= \sum_{t\geq0}\gamma^t \sum_{s\in\SC'}\sum_{a\in\A'}Pr(s_t=s)\pi(a|O(s))R(s, a)\\
&= \sum_{t\geq0}\gamma^t \sum_{s\in\SC'}\sum_{o\in\Omega}\sum_{a\in\A'}Pr(s_t=s|o_t=o)Pr(o_t=o)\pi(a|O(s))R'(s, a)\\
&=\sum_{t\geq0} \gamma^t\sum_{o\in\Omega}\sum_{a\in\A'}Pr(v_t=o)\pi(a|o)\sum_{s\in\SC'}p(s|o)R'(s, a)\\
&=\sum_{t\geq0} \gamma^t\sum_{o\in\Omega}\sum_{a\in\A'}Pr(v_t=o)\pi(a|o)R''(o, a)\\
&= J_{O}(\pi)
\end{align}
Thus reactive policies have the same policy return in the IBMDP and the OIBMDP. Since we know that an MDP always admits a deterministic policy as a solution, an optimal policy of the OIBMDP has a return $J(\pi^*)$, the return of the best deterministic reactive policy.
\newpage
\tableofcontents
\end{document}